\PassOptionsToPackage{dvipsnames,table,xcdraw}{xcolor}
\documentclass[]{fairmeta}
\usepackage{wrapfig}
\usepackage{tabularx}
\usepackage{textcomp}
\usepackage{stfloats}
\usepackage{url}
\usepackage{verbatim}
\usepackage{graphicx}
\usepackage{titlesec}
\usepackage{tocloft}
\usepackage{adjustbox}
\usepackage{multirow}
\usepackage{pifont}
\usepackage{tikz}
\usepackage{comment}
\usepackage{amsmath,amssymb} % define this before the line numbering.
\usepackage{booktabs} 
\usepackage{hyperref}
\usepackage{graphicx}    % 用于插入图片
\usepackage{subcaption} 
\usepackage{multirow} % 表格多行合并
\usepackage{booktabs} % 表格更好的线条
\usepackage{subcaption} % 支持 subtable 和 subfigure
\usepackage{booktabs}
\usepackage{enumitem}
\usepackage{graphicx}
\usepackage{amsmath}
\usepackage{amssymb}
\usepackage{pifont}
\usepackage{array}
\usepackage{multirow}
\usepackage{tablefootnote}
\usepackage{pifont}
\usepackage{tcolorbox}
\usepackage{caption}
\usepackage{float}
\usepackage[T1]{fontenc}
\usepackage[utf8]{inputenc}
\usepackage{microtype}
\RequirePackage{xspace}
\makeatletter
\DeclareRobustCommand\onedot{\futurelet\@let@token\@onedot}
\def\@onedot{\ifx\@let@token.\else.\null\fi\xspace}

\makeatother

\definecolor{adptorange}{RGB}{248, 205, 172}
\definecolor{cmpblue}{RGB}{189, 215, 238}
\definecolor{cmpblue}{RGB}{189, 215, 238}

\definecolor{our_red}{RGB}{232,157,160}
\definecolor{our_blue}{RGB}{136,206,230}
\definecolor{our_orange}{RGB}{246,200,168}
\definecolor{our_green}{RGB}{178,211,164}

\definecolor{attn_code0}{RGB}{247,215,200}
\definecolor{attn_code1}{RGB}{238,169,139}
\definecolor{mlp_code0}{RGB}{204,201,221}
\definecolor{mlp_code1}{RGB}{102,95,153}

\definecolor{token_blue}{RGB}{84, 120, 140}
\usepackage{pifont}       % \ding{xx}
\usepackage{bbding}       % \Checkmark,\XSolid,... (需要和pifont宏包共同使用)
\usepackage{fontawesome}
\usepackage{xspace}

\usepackage{float}

\newlength\savewidth

\newcolumntype{x}[1]{>{\centering\arraybackslash}p{#1pt}}
\newcolumntype{y}[1]{>{\raggedright\arraybackslash}p{#1pt}}
\newcolumntype{z}[1]{>{\raggedleft\arraybackslash}p{#1pt}}

\renewcommand{\paragraph}[1]{\vspace{1mm}\noindent\textbf{#1}}
\usepackage{colortbl}
\usepackage{xcolor}
\usepackage{wrapfig}

\renewcommand{\paragraph}[1]{\vspace{1.25mm}\noindent\textbf{#1}}

\usepackage{algorithm}
\usepackage{listings}

\definecolor{codeblue}{rgb}{0.25, 0.5, 0.5}
\definecolor{codekw}{rgb}{0.35, 0.35, 0.75}
\lstdefinestyle{Pytorch}{
    language = Python,
    backgroundcolor = \color{white},
    basicstyle = \fontsize{9pt}{8pt}\selectfont\ttfamily\bfseries,
    columns = fullflexible,
    aboveskip=1pt,
    belowskip=1pt,
    breaklines = true,
    captionpos = b,
    commentstyle = \color{codeblue},
    keywordstyle = \color{codekw},
}

\definecolor{green}{HTML}{009000}
\definecolor{red}{HTML}{ea4335}

\title{Benchmarking Multimodal Mathematical Reasoning with Explicit Visual Dependency}

\author[*1,2]{Zhikai Wang}
\author[*1,2]{Jiashuo Sun}
\author[1,3]{Wenqi Zhang}
\author[1,4]{Zhiqiang Hu}
\author[\dagger1,2]{Xin Li}
\author[1,2]{Fan Wang}
\author[1,2]{Deli Zhao}

\affiliation[1]{DAMO Academy, Alibaba Group}
\affiliation[2]{Hupan Lab\\}
\affiliation[3]{Zhejiang University}
\affiliation[4]{Singapore University of Technology and Design\\}

\contribution[*]{Equal contribution}
\contribution[\dagger]{Corresponding author}

\abstract{
Recent advancements in Large Vision-Language Models (LVLMs) have significantly enhanced their ability to integrate visual and linguistic information, achieving near-human proficiency in tasks like object recognition, captioning, and visual question answering. However, current benchmarks typically focus on knowledge-centric evaluations that assess domain-specific expertise, often neglecting the core ability to reason about fundamental mathematical elements and visual concepts. We identify a gap in evaluating elementary-level math problems, which rely on explicit visual dependencies-requiring models to discern, integrate, and reason across multiple images while incorporating commonsense knowledge, all o which are crucial for advancing toward broader AGI capabilities. To address this gap, we introduce \textsc{VCBench}, a comprehensive benchmark for multimodal mathematical reasoning with explicit visual dependencies. \textsc{VCBench} includes 1,720 problems across six cognitive domains, featuring 6,697 images (averaging 3.9 per question) to ensure multi-image reasoning. We evaluate 26 state-of-the-art LVLMs on \textsc{VCBench}, revealing substantial performance disparities, with even the top models unable to exceed 50\% accuracy. Our findings highlight the ongoing challenges in visual-mathematical integration and suggest avenues for future LVLM advancements.
}

\date{\today}

\begin{document}
\thispagestyle{firstheader}
\maketitle
\pagestyle{empty}

\begin{figure*}[t]
  \centering
  \includegraphics[width=\textwidth]{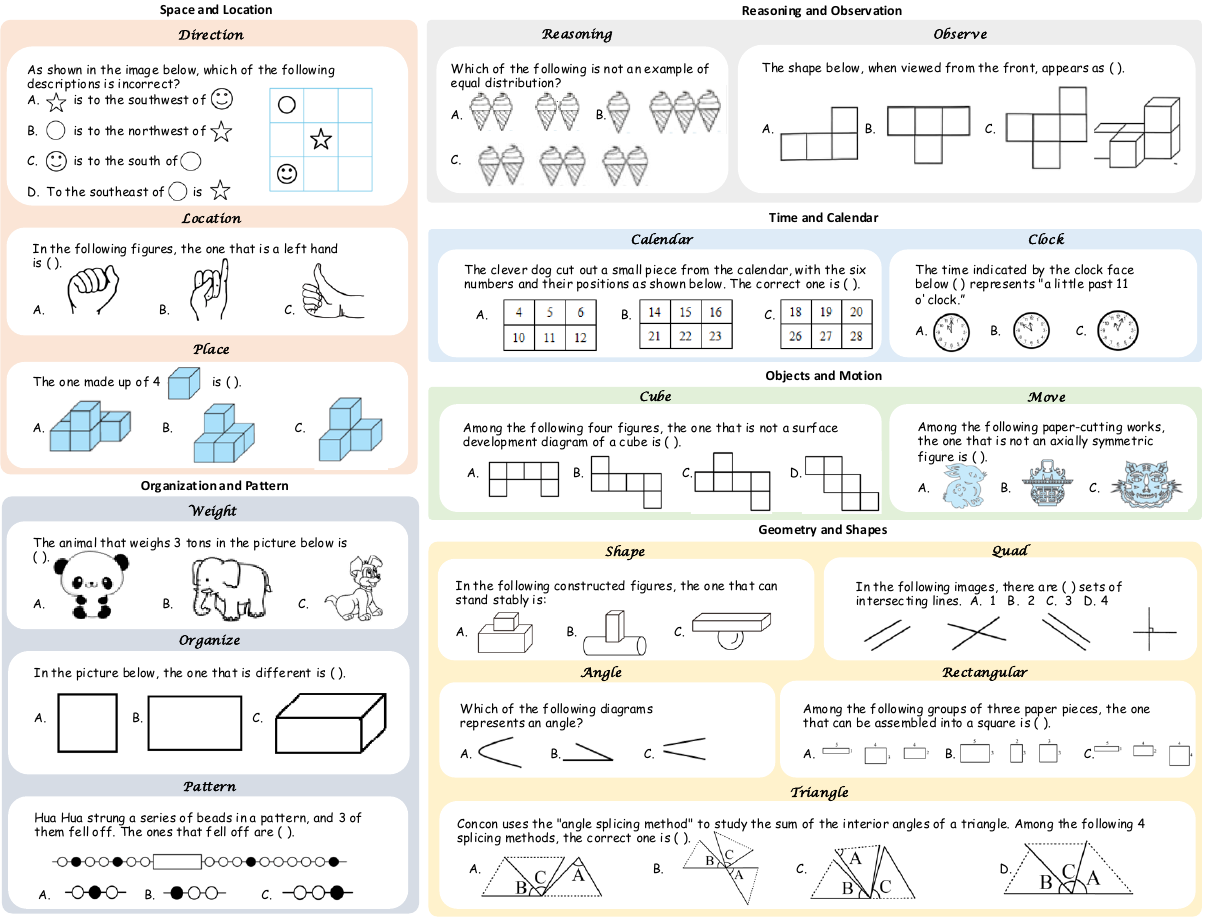}
\caption{Representative examples from the \textsc{VCBench}, showcasing diverse question types and categories including Space and Location (Direction, Location and Place), Reasoning and Observation (Reasoning and Observe), Time and Calendar (Calendar and Clock), Objects and Motion (Cube and Move), Organization and Pattern (Weight, Organize and Pattern), and Geometry and Shapes (Shape, Quad, Angle, Rectangular and Triangle).}
  \label{fig:pipeline}
\end{figure*}

\section{Introduction}
% 第一段，讲一下LVLM的发展，说明visual math的表现接近专家人类水平。

% 第二段，说明虽然有很多现有的visual math benchmark，但是他们侧重于。。。。但是，对于通往AGI，这种层次的评估远远不够。

% 第三段，我们发现，对小学水平的数学题来说，他们的特点是不需要很复杂的数学或几何层面的推理，但是具有明确的视觉依赖性。（这里可以讲一下孩童脑部发育，即目前他们不需要接受很复杂的推理知识，但是评估他们的能力却更需要他们对多个图像之间捕捉视觉特征和依赖，从而回答问题。）

% 第四段，因此，我们提出了\textsc{VCBench}，这是一个具有明确视觉依赖性的多模态数学推理的综合评估基准。 \textsc{\textsc{VCBench}} 是根据需要多幅图像才能解决的基本数学问题（1-6 年级）精心构建的，分为六个主要认知领域：时间和日历、空间和位置、几何和形状、物体和运动、推理和观察以及组织和模式。 \textsc{\textsc{VCBench}} 有 2,370 个问答对，包含 9,284 幅图像（平均每个问题 3.9 幅图像），确保模型必须跨视觉输入进行推理，而不是依赖于单幅图像的理解。

% 第五段我们进行了广泛的实验，评估了基准的 17 个不同任务类别中的 30 个最先进的 LVLM。我们发现，对正常人类在回答准确率接近满分的情况下，目前最先进的视觉模型准确率却达不到50%。

Recent advancements in Large Vision-Language Models (LVLMs) \cite{calude, gemini-2.0, openai2024gpt4ocard, qwen-vl-max} have made significant strides in bridging the gap between visual understanding and language processing. These models have achieved remarkable performance across a range of tasks, demonstrating near-expert human-level proficiency in domains such as object recognition, caption generation, and visual question answering \cite{coco, vqa}. Among the various domains explored, LVLMs have shown particular promise in tasks that require both visual and linguistic reasoning, making them increasingly relevant for real-world applications.

While many visual mathematics benchmarks, such as MathVista \cite{lu2023mathvista} and MathVision \cite{math_vision}, focus on knowledge-centric evaluations that assess domain-specific mathematical or geometric expertise, they often fail to evaluate a model's core ability to perceive and reason about fundamental mathematical elements and visual concepts. Moreover, these knowledge-centric evaluations are easily influenced by the pre-existing knowledge embedded in large language models, which may obscure true reasoning capabilities. To advance towards Artificial General Intelligence (AGI), a more holistic approach to multi-modal reasoning is needed—one that goes beyond task-specific benchmarks and better captures generalizable cognitive abilities.

In this context, we identify a gap in the evaluation of models on elementary-level math problems \cite{gsm8k, cmath}. These problems, typically at the elementary school level, do not require complex mathematical or geometric reasoning but rely heavily on explicit visual dependencies—the ability to discern and integrate visual features across images and understand how different visual elements relate to one another to solve problems. This mirrors the cognitive development of children, who, at a young age, rely on similar skills to solve problems despite not yet possessing advanced reasoning abilities. Understanding and modeling this form of reasoning is crucial, as it represents a fundamental cognitive ability essential for advancing toward broader AGI capabilities.

To address this gap, we introduce \textsc{VCBench}, a comprehensive benchmark designed to assess multimodal mathematical reasoning tasks with explicit visual dependencies. Specifically targeting elementary-level math problems (grades 1–6), \textsc{VCBench} focuses on tasks that require reasoning across multiple images to derive solutions. As shown in Figure \ref{fig:pipeline}, it covers six key cognitive domains: Time and Calendar, Spatial and Positional Awareness, Geometry and Shapes, Objects and Motion, Reasoning and Observation, and Organization and Patterns. It also evaluates five distinct competencies: temporal reasoning, geometric reasoning, logical reasoning, spatial reasoning, and pattern recognition. These competencies span a broad spectrum, from basic temporal and spatial understanding to more advanced geometric and logical reasoning, providing a thorough evaluation of multimodal model performance. Comprising 1,720 question-answer pairs and 6,697 images (averaging 3.9 images per question), \textsc{VCBench} ensures models must reason across multiple visual inputs, rather than relying on single-image comprehension. With this holistic framework, \textsc{VCBench} serves as a valuable resource for advancing research in multimodal mathematical reasoning.

In our extensive experimental evaluation, we assessed 26 state-of-the-art LVLMs across 17 distinct task categories within \textsc{VCBench}. Despite achieving near-perfect accuracy on normal human-level performance, the best-performing visual models were unable to exceed 50\% accuracy. Many of these state-of-the-art models exhibited a notable lack of pattern recognition in images, especially when it came to reasoning tasks that required integrating visual cues across multiple images. Interestingly, we observed that these same tasks could be easily answered by normal human. This highlights a significant gap in current benchmarks, which fail to adequately assess vision-centric mathematical reasoning abilities.

We make several key contributions with \textsc{VCBench}:

Unlike existing benchmarks that focus on knowledge-centric evaluations, we emphasize vision-centric assessments. \textsc{VCBench} targets problems that do not require specialized knowledge but instead rely on the common perceptual reasoning of mathematical images and concepts. This approach aligns with the way children learn—first mastering visual reasoning and only later acquiring domain-specific knowledge.

\textsc{VCBench} is designed around multi-image tasks, with each question containing an average of 3.9 images. This requirement challenges models to explicitly integrate visual cues across multiple images and reason about how they interact, which better reflects real-world scenarios where information is often distributed across multiple visual inputs.

Our benchmark provides a holistic evaluation of various visual reasoning capabilities, such as temporal reasoning, spatial understanding, and pattern recognition. While these tasks may seem simple to children, they represent fundamental reasoning abilities that LVLMs often struggle with. Our experiments demonstrate that tasks considered easy for children—such as identifying time sequences or spatial relationships—prove challenging for state-of-the-art LVLMs, highlighting the gaps in current multimodal reasoning capabilities.

\begin{figure}[t]
  \centering
  \begin{subfigure}[t]{0.48\textwidth}
    \centering
    \includegraphics[width=\textwidth]{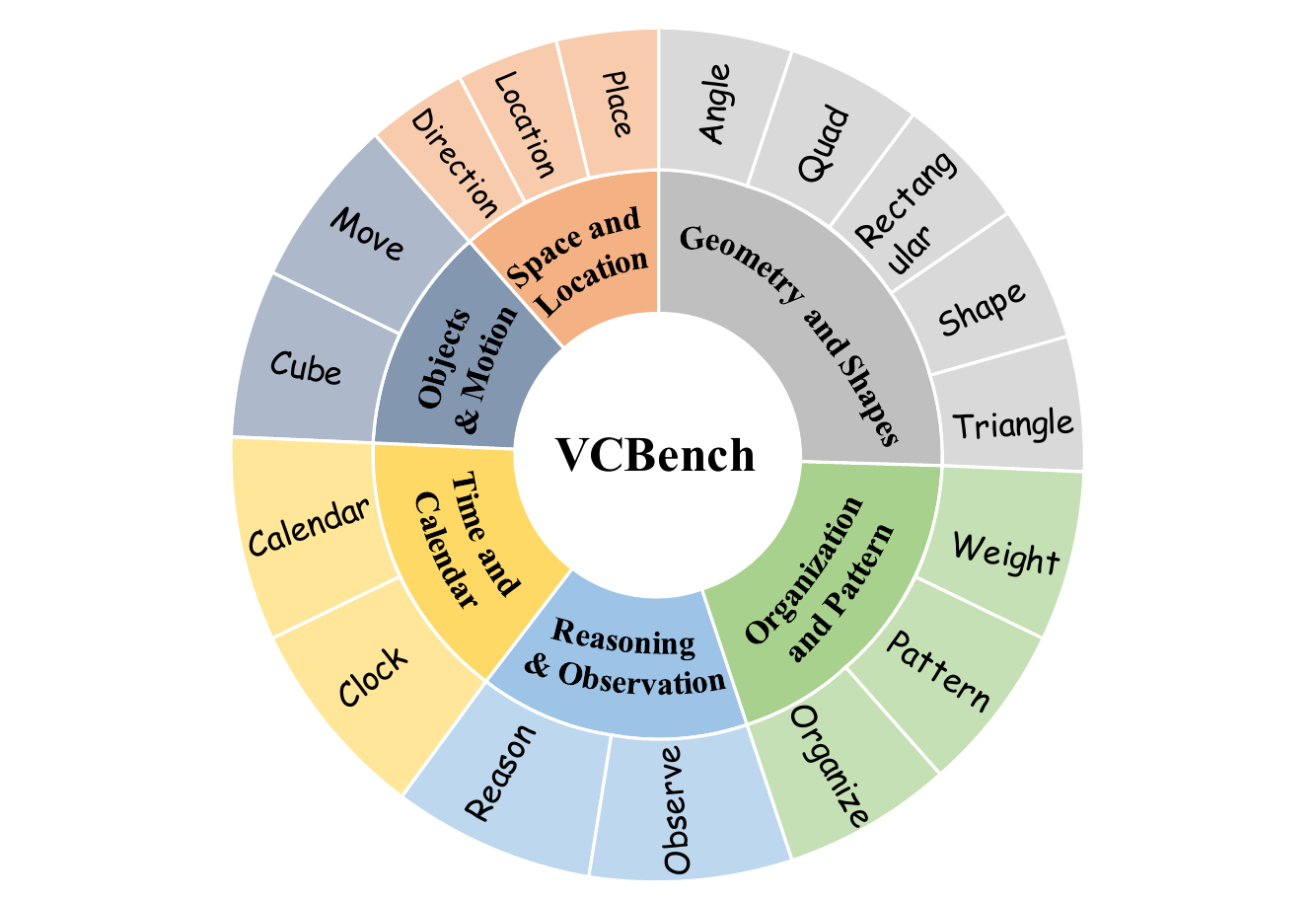}
    \caption{}
    \label{fig:vc_bench}
  \end{subfigure}
  \hfill
  \begin{subfigure}[t]{0.48\textwidth}
    \centering
    \includegraphics[width=\textwidth]{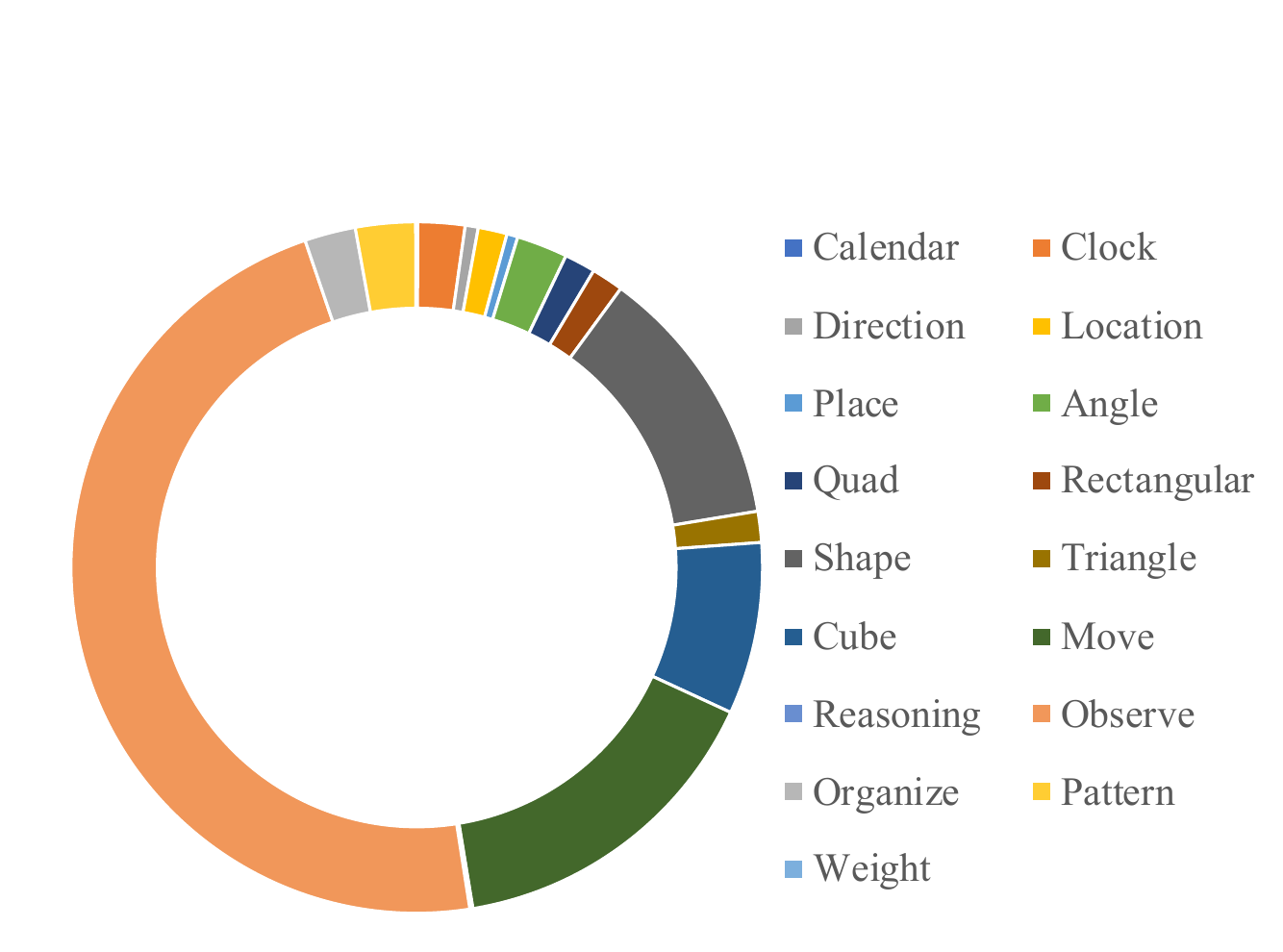}
    \caption{}
    \label{fig:distribution}
  \end{subfigure}
  \caption{(a)~Overview of the \textsc{VCBench} dataset structure, highlighting its six main categories and associated subcategories, designed to assess multimodal reasoning capabilities of LVLMs. (b)~Distribution of question types in the \textsc{VCBench}, illustrating the relative frequency across different visual reasoning subcategories}
  \label{fig:vcbench_overview}
\end{figure}

\begin{table}[t]
\small
\begin{center}
\caption{Comprehensive Statistics of the \textsc{\textsc{VCBench}} Dataset, Including Detailed Breakdown of Question-Image Pairs, Image Distribution, and Question Length Metrics.}
\resizebox{0.35\textwidth}{!}{%
\begin{tabular}{lr}
\toprule
Examples (Q\&A pairs) & 1,720\\
Images & 6,697 \\
Avg. images per question & 3.9\\
Avg. question length & 136.2\\
Max. \# images in question & 18 \\
Min. \# images in question & 2 \\
\bottomrule
\end{tabular}
}
\label{table:gen_stats}
\end{center}
\end{table}

\begin{figure}[t]
  \centering
  \includegraphics[width=0.58\textwidth]{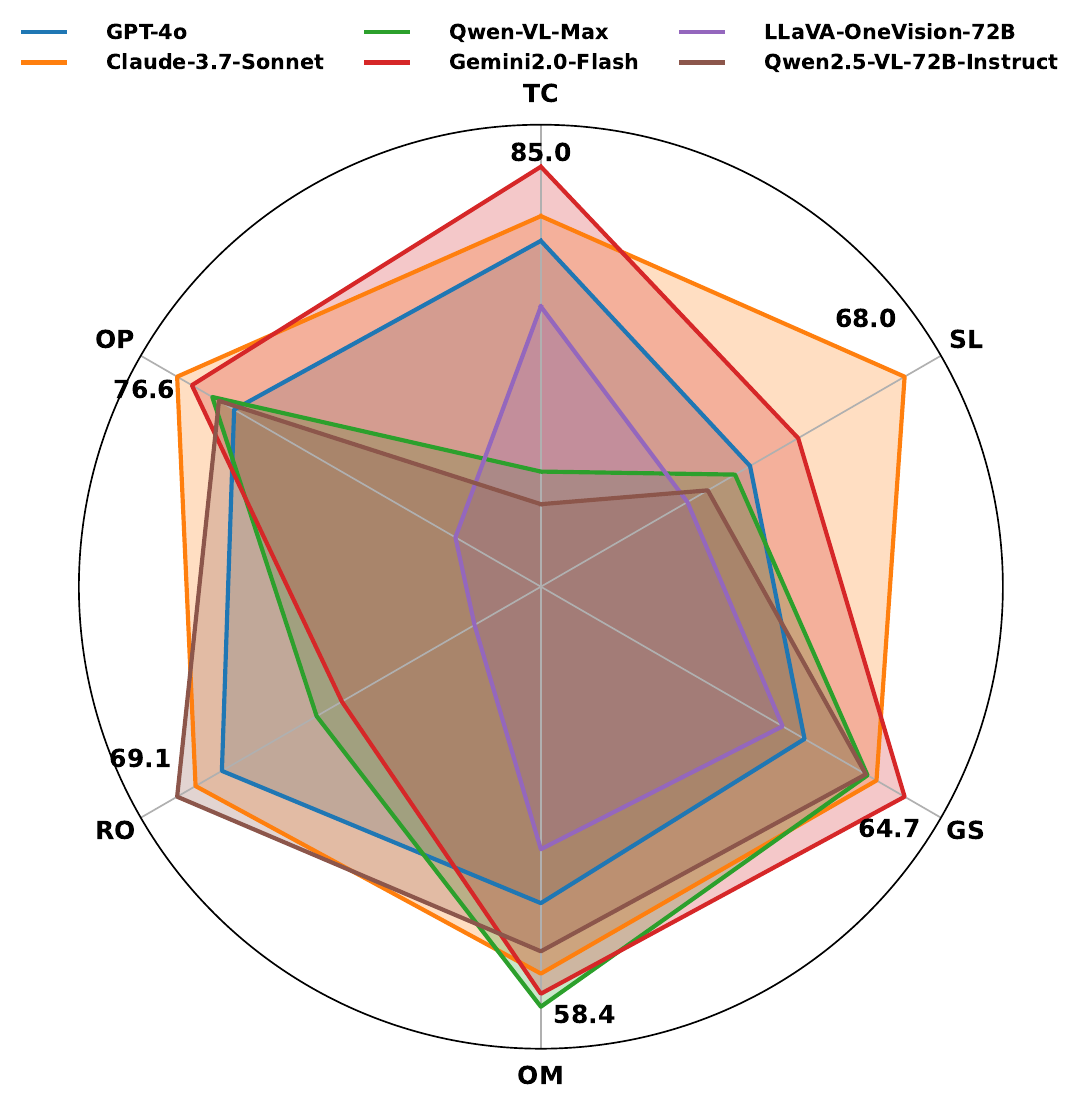}
  \caption{Comparative performance (\%) of six various prominent LVLMs across six categories: Time and Calendar (TC), Space and Location (SL), Geometry and Shapes (GS), Objects and Motion (OM), Reasoning and Observation (RO), and Organization and Pattern (OP).}
  \label{fig:radar}
\end{figure}

\begin{figure*}[t]
  \centering
  \includegraphics[width=1.0\textwidth]{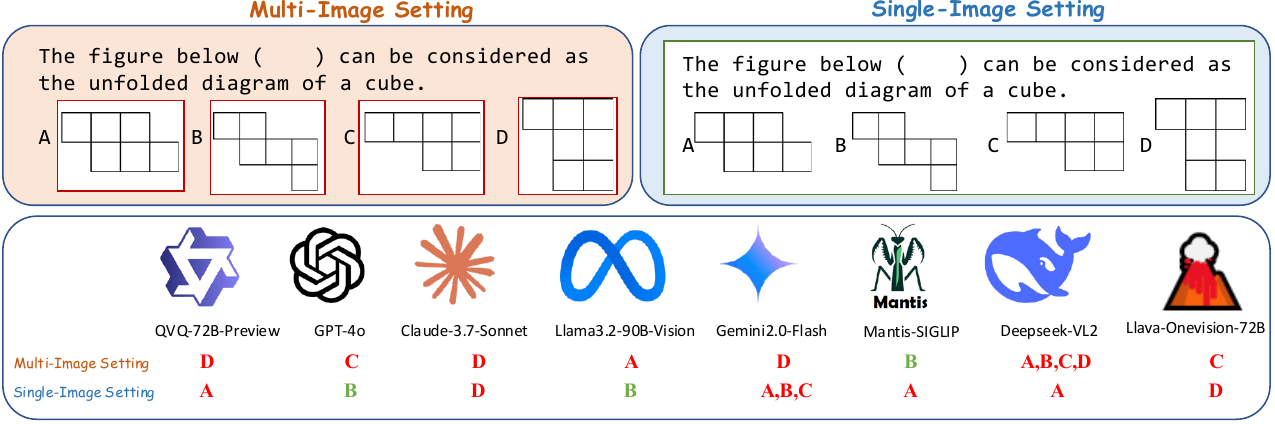}
  \caption{Comparative evaluation of various LVLMs under Multi-Image and Single-Image settings for the same question. The letters (A, B, C, D) indicate models' predictions, with correct answers marked in \textcolor{ForestGreen}{green} and incorrect answers in \textcolor{red}{red}.}
  \label{fig:overview}
\end{figure*}

\section{Related Work}

\paragraph{Large Vision-Language Models.}
Large Vision-Language Models (LVLMs) have significantly advanced the integration of vision and language, demonstrating strong performance in tasks such as image captioning, visual question answering (VQA), and complex multimodal reasoning \cite{wang2024e, wu2023a}. Recent developments, such as Gemini-2.0 \cite{gemini-2.0}, QVQ \cite{qvq-72b-preview}, and Calude-3.7-Sonnet \cite{calude}, showcase emergent abilities in cross-modal instruction-following and chain-of-thought reasoning.

Despite these advancements, quantitatively evaluating LVLMs, particularly in visual mathematical reasoning, remains challenging. Existing benchmarks like MathVista \cite{lu2023mathvista}, MathBench \cite{liu2024b}, and Math-LLMs \cite{liu2023b} typically assess models within narrow domains, such as arithmetic word problems or geometry-based visual environments. Consequently, these benchmarks primarily measure foundational skills like geometric or spatial reasoning, limiting their capacity to comprehensively evaluate broader cognitive integration and reasoning abilities. To address this limitation, we introduce \textsc{\textsc{VCBench}}, a systematic evaluation framework designed to rigorously assess LVLMs performance across diverse multimodal mathematical reasoning tasks with explicit visual dependencies.

\paragraph{Visual Mathematical Reasoning.}
Mathematical reasoning is a core cognitive ability increasingly explored within the context of LVLMs research \cite{hendrycks2021math, cobbe2021gsm8k}. While earlier benchmarks such as GSM8K \cite{cobbe2021gsm8k} and MATH \cite{hendrycks2021math} primarily focused on text-based mathematical problems, recent research has expanded toward visual mathematical reasoning, incorporating diagrams, charts, and geometry-based problem-solving \cite{wang2024e, yang2024b}.

Multimodal mathematical reasoning requires LVLMs to simultaneously integrate visual perception and logical reasoning, presenting a greater challenge compared to purely textual problems. Recent benchmarks like MathVista \cite{lu2023mathvista} and MathGLM-Vision \cite{yang2024b} have advanced evaluation efforts but still suffer from issues including ambiguous annotations, dependency on GPT-based scoring methods, and limited evaluation of generalizable cognitive abilities \cite{yan2024a}.

To overcome these challenges, we propose\textsc{\textsc{VCBench}}, a comprehensive benchmark explicitly designed for multimodal mathematical reasoning with visual dependencies. \textsc{\textsc{VCBench}} encompasses 17 distinct subtasks, systematically assessing foundational cognitive skills such as temporal reasoning, logical reasoning, spatial reasoning, geometric reasoning, and pattern recognition. By standardizing task instructions and employing a multiple-choice evaluation format, \textsc{\textsc{VCBench}} provides objective, reproducible evaluations, offering deeper insights into the strengths and limitations of current LVLMs.

\section{\textsc{\textsc{VCBench}}}

\subsection{Benchmark Construction}
For \textsc{\textsc{VCBench}}, we employed a systematic approach to collect high-quality multimodal mathematical reasoning problems that explicitly require visual reasoning. We started by examining Chinese elementary school mathematics textbooks from grades 1-6, manually filtering for problems that contained at least two images. The benchmark prioritizes vision-centric evaluation through perceptual reasoning tasks that avoid specialized knowledge, while simultaneously challenging models to implicitly integrate and synthesize visual information across multiple images - a critical capability for real-world applications where understanding emerges from connecting disparate visual cues.
During our selection process, we enforced strict criteria to ensure quality and consistency. We only retained problems with unique, unambiguous answers to facilitate objective evaluation. After the initial collection phase, we utilized large language models to translate all problems into English (the specific prompts used are available in the Appendix), followed by rigorous human verification to maintain translation accuracy.
The human verification process also served as a secondary filtering mechanism, where we eliminated problems containing non-English content in images, as well as those with unclear visual elements or ambiguous instructions. This meticulous curation process ensured that our benchmark evaluates genuine reasoning abilities rather than testing models on their capacity to handle poorly defined problems.
Through this methodology, we assembled our final collection of problems that encompass various mathematical domains while maintaining consistent quality standards.

\subsection{Benchmark Statistics}

\textsc{\textsc{VCBench}} comprises a diverse collection of multimodal mathematical reasoning problems, carefully organized into \textbf{six} major categories to provide comprehensive coverage of different cognitive dimensions. As shown in Table \ref{table:gen_stats}, our benchmark contains 1,720 question-answer pairs featuring a total of 6,697 images. The average question includes 3.9 images, with some complex problems containing up to 18 images, while the minimum is 2 images per question.

To systematically evaluate different reasoning capabilities, we categorized our problems into six major domains:

\begin{itemize}[left=0pt]
    \item \textbf{Time and Calendar}: Problems testing temporal reasoning across two subcategories (\textit{Calendar} and \textit{Clock}) that require understanding time intervals, and calendar-based calculations.
    
    \item \textbf{Space and Location}: Challenges focused on spatial reasoning across three subcategories (\textit{Direction}, \textit{Location}, and \textit{Place}) that assess understanding of relative positions, directions, and spatial relationships.
    
    \item \textbf{Geometry and Shapes}: Problems spanning five subcategories (\textit{Angle}, \textit{Quad}, \textit{Rectangular}, \textit{Shape}, and \textit{Triangle}) that test fundamental geometric comprehension from basic shape recognition to more complex property analysis.
    
    \item \textbf{Objects and Motion}: Tasks in two subcategories (\textit{Cube} and \textit{Move}) that evaluate the understanding of three-dimensional objects and motion transformations.
    
    \item \textbf{Reasoning and Observation}: Problems in two subcategories (\textit{Reasoning} and \textit{Observe}) designed to test logical reasoning and careful visual observation skills.
    
    \item \textbf{Organization and Pattern}: Challenges across three subcategories (\textit{Organize}, \textit{Pattern}, and \textit{Weight}) that assess pattern recognition, sequencing, and organizational logic.
\end{itemize}

This categorization allows for a granular assessment of model performance across specific cognitive abilities required for mathematical reasoning. The vocabulary in our benchmark is deliberately controlled to 2,312 unique words, ensuring that performance differences reflect reasoning capabilities rather than linguistic complexity. With an average question length of 136.2 characters, our problems are concise yet sufficiently detailed to communicate the necessary context for solving each problem.

\begin{table*}[t]
\centering
\caption{Performance of various vision-language models (Close-Source, Open-Source, and Math Specialist categories) on a Multi-image setting across multiple tasks, including Time and Calendar, Space and Location, Geometry and Shapes, Objects and Motion, Reasoning and Observation, and Organization and Pattern.}
\resizebox{\textwidth}{!}{
\begin{tabular}{lcccccccccccccccccc}
\toprule
\multirow{2}{*}{\textbf{Models}} & \multicolumn{2}{c}{\textbf{Time and Calendar}} & \multicolumn{3}{c}{\textbf{Space and Location}} & \multicolumn{5}{c}{\textbf{Geometry and Shapes}} & \multicolumn{2}{c}{\textbf{Objects and Motion}} & \multicolumn{2}{c}{\textbf{Reasoning and Observation}} & \multicolumn{3}{c}{\textbf{Organization and Pattern}} & \multirow{2}{*}{\textbf{Avg.}} \\
                                & \textbf{Calender} & \textbf{Clock} & \textbf{Direction} & \textbf{Location} & \textbf{Place} & \textbf{Angle} & \textbf{Quad} & \textbf{Rectangular} & \textbf{Shape} & \textbf{Triangle} & \textbf{Cube} & \textbf{Move} & \textbf{Reasoning} & \textbf{Observe} & \textbf{Organize} & \textbf{Pattern} & \textbf{Weight} & \\
\midrule
Random Guess            & \cellcolor{blue!10}33.33 & \cellcolor{blue!10}32.78 & \cellcolor{green!10}25.00 & \cellcolor{green!10}29.81 & \cellcolor{green!10}33.33 & \cellcolor{yellow!10}31.00 & \cellcolor{yellow!10}27.63 & \cellcolor{yellow!10}29.17 & \cellcolor{yellow!10}31.84 & \cellcolor{yellow!10}29.01 & \cellcolor{orange!10}28.37 & \cellcolor{orange!10}29.35 & \cellcolor{teal!10}33.33 & \cellcolor{teal!10}29.41 & \cellcolor{red!10}30.17 & \cellcolor{red!10}31.32 & \cellcolor{red!10}33.33 & \cellcolor{gray!10}29.83 \\ 
Human                  & \cellcolor{blue!10}100.00 & \cellcolor{blue!10}96.00 & \cellcolor{green!10}100.00 & \cellcolor{green!10}93.85 & \cellcolor{green!10}96.67 & \cellcolor{yellow!10}95.60 & \cellcolor{yellow!10}96.84 & \cellcolor{yellow!10}95.00 & \cellcolor{yellow!10}94.02 & \cellcolor{yellow!10}94.07 & \cellcolor{orange!10}97.67 & \cellcolor{orange!10}94.63 & \cellcolor{teal!10}100.00 & \cellcolor{teal!10}93.59 & \cellcolor{red!10}93.20 & \cellcolor{red!10}95.52 & \cellcolor{red!10}100.00 & \cellcolor{gray!10}93.30 \\ 
\multicolumn{19}{c}{\cellcolor{gray!20}\textit{\textbf{Close-Source Models}}} \\
GPT-4o-mini \cite{openai2024gpt4ocard} & \cellcolor{blue!10}80.00 & \cellcolor{blue!10}60.66 & \cellcolor{green!10}0.00 & \cellcolor{green!10}38.46 & \cellcolor{green!10}53.33 & \cellcolor{yellow!10}38.40 & \cellcolor{yellow!10}21.05 & \cellcolor{yellow!10}53.57 & \cellcolor{yellow!10}37.99 & \cellcolor{yellow!10}55.56 & \cellcolor{orange!10}32.19 & \cellcolor{orange!10}38.24 & \cellcolor{teal!10}0.00 & \cellcolor{teal!10}28.68 & \cellcolor{red!10}60.00 & \cellcolor{red!10}41.38 & \cellcolor{red!10}100.00 & \cellcolor{gray!10}34.88 \\
GPT-4o \cite{openai2024gpt4ocard} & \cellcolor{blue!10}100.00 & \cellcolor{blue!10}40.00 & \cellcolor{green!10}20.00 & \cellcolor{green!10}30.77 & \cellcolor{green!10}66.67 & \cellcolor{yellow!10}46.00 & \cellcolor{yellow!10}57.89 & \cellcolor{yellow!10}28.57 & \cellcolor{yellow!10}50.22 & \cellcolor{yellow!10}51.85 & \cellcolor{orange!10}37.67 & \cellcolor{orange!10}50.37 & \cellcolor{teal!10}90.00 & \cellcolor{teal!10}31.27 & \cellcolor{red!10}76.00 & \cellcolor{red!10}37.93 & \cellcolor{red!10}80.00 & \cellcolor{gray!10}40.29 \\
Claude-3.7-Sonnet \cite{calude} & \cellcolor{blue!10}100.00 & \cellcolor{blue!10}50.00 & \cellcolor{green!10}100.00 & \cellcolor{green!10}53.85 & \cellcolor{green!10}50.00 & \cellcolor{yellow!10}58.00 & \cellcolor{yellow!10}63.16 & \cellcolor{yellow!10}57.14 & \cellcolor{yellow!10}60.70 & \cellcolor{yellow!10}59.26 & \cellcolor{orange!10}40.41 & \cellcolor{orange!10}67.28 & \cellcolor{teal!10}100.00 & \cellcolor{teal!10}31.27 & \cellcolor{red!10}76.40 & \cellcolor{red!10}53.45 & \cellcolor{red!10}100.00 & \cellcolor{gray!10}46.63 \\
Qwen-VL-Max \cite{qwen-vl-max} & \cellcolor{blue!10}0.00 & \cellcolor{blue!10}46.67 & \cellcolor{green!10}0.00 & \cellcolor{green!10}42.31 & \cellcolor{green!10}66.67 & \cellcolor{yellow!10}74.00 & \cellcolor{yellow!10}52.63 & \cellcolor{yellow!10}42.86 & \cellcolor{yellow!10}54.15 & \cellcolor{yellow!10}66.67 & \cellcolor{orange!10}56.16 & \cellcolor{orange!10}60.66 & \cellcolor{teal!10}50.00 & \cellcolor{teal!10}35.27 & \cellcolor{red!10}68.00 & \cellcolor{red!10}39.66 & \cellcolor{red!10}100.00 & \cellcolor{gray!10}47.03 \\
Gemini2.0-Flash \cite{gemini-2.0} & \cellcolor{blue!10}100.00 & \cellcolor{blue!10}70.00 & \cellcolor{green!10}20.00 & \cellcolor{green!10}57.69 & \cellcolor{green!10}66.67 & \cellcolor{yellow!10}70.00 & \cellcolor{yellow!10}68.42 & \cellcolor{yellow!10}53.57 & \cellcolor{yellow!10}61.14 & \cellcolor{yellow!10}70.37 & \cellcolor{orange!10}44.52 & \cellcolor{orange!10}68.75 & \cellcolor{teal!10}40.00 & \cellcolor{teal!10}35.53 & \cellcolor{red!10}74.00 & \cellcolor{red!10}46.55 & \cellcolor{red!10}100.00 & \cellcolor{gray!10}49.77 \\

\multicolumn{19}{c}{\cellcolor{gray!20}\textit{\textbf{Open-Source Models}}} \\
Emu2-Chat \cite{emu2} & \cellcolor{blue!10}0.00 & \cellcolor{blue!10}13.33 & \cellcolor{green!10}0.00 & \cellcolor{green!10}3.85 & \cellcolor{green!10}0.00 & \cellcolor{yellow!10}4.00 & \cellcolor{yellow!10}10.53 & \cellcolor{yellow!10}10.71 & \cellcolor{yellow!10}12.66 & \cellcolor{yellow!10}3.70 & \cellcolor{orange!10}8.90 & \cellcolor{orange!10}6.99 & \cellcolor{teal!10}0.00 & \cellcolor{teal!10}3.62 & \cellcolor{red!10}0.00 & \cellcolor{red!10}3.45 & \cellcolor{red!10}20.00 & \cellcolor{gray!10}6.05 \\
Idefics3-8B \cite{idefics} & \cellcolor{blue!10}0.00 & \cellcolor{blue!10}3.33 & \cellcolor{green!10}20.00 & \cellcolor{green!10}15.38 & \cellcolor{green!10}33.33 & \cellcolor{yellow!10}11.60 & \cellcolor{yellow!10}10.53 & \cellcolor{yellow!10}17.86 & \cellcolor{yellow!10}23.14 & \cellcolor{yellow!10}3.70 & \cellcolor{orange!10}9.59 & \cellcolor{orange!10}16.91 & \cellcolor{teal!10}0.00 & \cellcolor{teal!10}9.69 & \cellcolor{red!10}8.40 & \cellcolor{red!10}15.52 & \cellcolor{red!10}0.00 & \cellcolor{gray!10}12.91 \\
DeepSeek-VL2 \cite{deepseek-vl-2} & \cellcolor{blue!10}0.00 & \cellcolor{blue!10}23.33 & \cellcolor{green!10}0.00 & \cellcolor{green!10}23.08 & \cellcolor{green!10}16.67 & \cellcolor{yellow!10}14.00 & \cellcolor{yellow!10}10.53 & \cellcolor{yellow!10}14.29 & \cellcolor{yellow!10}29.69 & \cellcolor{yellow!10}14.81 & \cellcolor{orange!10}6.85 & \cellcolor{orange!10}18.38 & \cellcolor{teal!10}10.00 & \cellcolor{teal!10}9.43 & \cellcolor{red!10}44.00 & \cellcolor{red!10}20.69 & \cellcolor{red!10}0.00 & \cellcolor{gray!10}15.47 \\
Phi-3.5-vision-instruct \cite{phi-3.5} & \cellcolor{blue!10}0.00 & \cellcolor{blue!10}23.33 & \cellcolor{green!10}100.00 & \cellcolor{green!10}19.23 & \cellcolor{green!10}66.67 & \cellcolor{yellow!10}16.00 & \cellcolor{yellow!10}15.79 & \cellcolor{yellow!10}28.57 & \cellcolor{yellow!10}27.07 & \cellcolor{yellow!10}33.33 & \cellcolor{orange!10}22.60 & \cellcolor{orange!10}22.79 & \cellcolor{teal!10}0.00 & \cellcolor{teal!10}21.32 & \cellcolor{red!10}34.00 & \cellcolor{red!10}12.07 & \cellcolor{red!10}0.00 & \cellcolor{gray!10}22.73 \\
InternVL2.5-8B \cite{internvl-2.5} & \cellcolor{blue!10}0.00 & \cellcolor{blue!10}33.33 & \cellcolor{green!10}0.00 & \cellcolor{green!10}34.62 & \cellcolor{green!10}50.00 & \cellcolor{yellow!10}34.00 & \cellcolor{yellow!10}31.58 & \cellcolor{yellow!10}50.00 & \cellcolor{yellow!10}35.81 & \cellcolor{yellow!10}37.04 & \cellcolor{orange!10}23.29 & \cellcolor{orange!10}25.74 & \cellcolor{teal!10}0.00 & \cellcolor{teal!10}18.99 & \cellcolor{red!10}38.00 & \cellcolor{red!10}6.90 & \cellcolor{red!10}0.00 & \cellcolor{gray!10}24.71 \\ 
Llama-3.2-90B-Vision-Instruct \cite{meta2024llama32} & \cellcolor{blue!10}20.00 & \cellcolor{blue!10}24.67 & \cellcolor{green!10}100.00 & \cellcolor{green!10}11.54 & \cellcolor{green!10}16.67 & \cellcolor{yellow!10}26.40 & \cellcolor{yellow!10}31.58 & \cellcolor{yellow!10}32.14 & \cellcolor{yellow!10}26.20 & \cellcolor{yellow!10}22.22 & \cellcolor{orange!10}27.40 & \cellcolor{orange!10}25.37 & \cellcolor{teal!10}0.00 & \cellcolor{teal!10}25.58 & \cellcolor{red!10}12.00 & \cellcolor{red!10}29.31 & \cellcolor{red!10}20.00 & \cellcolor{gray!10}25.41 \\
Qwen2.5-VL-7B-Instruct \cite{qwen-2.5-vl} & \cellcolor{blue!10}100.00 & \cellcolor{blue!10}13.33 & \cellcolor{green!10}0.00 & \cellcolor{green!10}19.23 & \cellcolor{green!10}50.00 & \cellcolor{yellow!10}20.00 & \cellcolor{yellow!10}31.58 & \cellcolor{yellow!10}25.00 & \cellcolor{yellow!10}30.13 & \cellcolor{yellow!10}51.85 & \cellcolor{orange!10}32.19 & \cellcolor{orange!10}40.81 & \cellcolor{teal!10}0.00 & \cellcolor{teal!10}25.19 & \cellcolor{red!10}30.00 & \cellcolor{red!10}27.59 & \cellcolor{red!10}0.00 & \cellcolor{gray!10}29.24 \\
Mantis-CLIP \cite{mantis} & \cellcolor{blue!10}0.00 & \cellcolor{blue!10}30.00 & \cellcolor{green!10}80.00 & \cellcolor{green!10}50.00 & \cellcolor{green!10}66.67 & \cellcolor{yellow!10}14.00 & \cellcolor{yellow!10}15.79 & \cellcolor{yellow!10}35.71 & \cellcolor{yellow!10}38.43 & \cellcolor{yellow!10}37.04 & \cellcolor{orange!10}19.86 & \cellcolor{orange!10}32.35 & \cellcolor{teal!10}40.00 & \cellcolor{teal!10}28.04 & \cellcolor{red!10}52.40 & \cellcolor{red!10}22.41 & \cellcolor{red!10}100.00 & \cellcolor{gray!10}30.23 \\
Mistral-Small-3.1-24B-Instruct \cite{mistral} & \cellcolor{blue!10}20.00 & \cellcolor{blue!10}40.00 & \cellcolor{green!10}0.00 & \cellcolor{green!10}30.77 & \cellcolor{green!10}30.00 & \cellcolor{yellow!10}38.00 & \cellcolor{yellow!10}31.58 & \cellcolor{yellow!10}35.71 & \cellcolor{yellow!10}29.26 & \cellcolor{yellow!10}51.85 & \cellcolor{orange!10}30.82 & \cellcolor{orange!10}31.62 & \cellcolor{teal!10}50.00 & \cellcolor{teal!10}29.59 & \cellcolor{red!10}38.00 & \cellcolor{red!10}34.48 & \cellcolor{red!10}20.00 & \cellcolor{gray!10}31.34 \\
Kimi-VL-A3B-Thinking\cite{kimiteam2025kimivltechnicalreport} & \cellcolor{blue!10}100.00 & \cellcolor{blue!10}26.67 & \cellcolor{green!10}100.00 & \cellcolor{green!10}30.77 & \cellcolor{green!10}33.33 & \cellcolor{yellow!10}48.00 & \cellcolor{yellow!10}36.84 & \cellcolor{yellow!10}28.57 & \cellcolor{yellow!10}49.78 & \cellcolor{yellow!10}33.33 & \cellcolor{orange!10}30.14 & \cellcolor{orange!10}41.91 & \cellcolor{teal!10}0.00 & \cellcolor{teal!10}25.32 & \cellcolor{red!10}68.00 & \cellcolor{red!10}27.59 & \cellcolor{red!10}100.00 & \cellcolor{gray!10}34.13 \\
LLaVA-Interleave-7B \cite{li2024llavanextinterleavetacklingmultiimagevideo} & \cellcolor{blue!10}0.00 & \cellcolor{blue!10}36.67 & \cellcolor{green!10}20.00 & \cellcolor{green!10}19.23 & \cellcolor{green!10}83.33 & \cellcolor{yellow!10}46.00 & \cellcolor{yellow!10}26.32 & \cellcolor{yellow!10}57.14 & \cellcolor{yellow!10}39.74 & \cellcolor{yellow!10}29.63 & \cellcolor{orange!10}30.82 & \cellcolor{orange!10}33.46 & \cellcolor{teal!10}50.00 & \cellcolor{teal!10}33.46 & \cellcolor{red!10}62.00 & \cellcolor{red!10}31.03 & \cellcolor{red!10}100.00 & \cellcolor{gray!10}35.47 \\
LLaVA-OneVision-7B \cite{llava-onevisoin} & \cellcolor{blue!10}0.00 & \cellcolor{blue!10}40.00 & \cellcolor{green!10}0.00 & \cellcolor{green!10}11.54 & \cellcolor{green!10}83.33 & \cellcolor{yellow!10}44.00 & \cellcolor{yellow!10}36.84 & \cellcolor{yellow!10}32.14 & \cellcolor{yellow!10}37.99 & \cellcolor{yellow!10}48.15 & \cellcolor{orange!10}30.82 & \cellcolor{orange!10}46.69 & \cellcolor{teal!10}50.00 & \cellcolor{teal!10}32.56 & \cellcolor{red!10}58.00 & \cellcolor{red!10}29.31 & \cellcolor{red!10}100.00 & \cellcolor{gray!10}36.63 \\
Kimi-VL-A3B-Instruct\cite{kimiteam2025kimivltechnicalreport} & \cellcolor{blue!10}0.00 & \cellcolor{blue!10}46.67 & \cellcolor{green!10}0.00 & \cellcolor{green!10}30.77 & \cellcolor{green!10}83.33 & \cellcolor{yellow!10}44.00 & \cellcolor{yellow!10}47.37 & \cellcolor{yellow!10}39.29 & \cellcolor{yellow!10}43.23 & \cellcolor{yellow!10}33.33 & \cellcolor{orange!10}34.93 & \cellcolor{orange!10}44.49 & \cellcolor{teal!10}50.00 & \cellcolor{teal!10}31.31 & \cellcolor{red!10}58.00 & \cellcolor{red!10}36.21 & \cellcolor{red!10}0.00 & \cellcolor{gray!10}37.33 \\
InternVL2.5-78B \cite{internvl-2.5} & \cellcolor{blue!10}20.00 & \cellcolor{blue!10}31.33 & \cellcolor{green!10}100.00 & \cellcolor{green!10}42.31 & \cellcolor{green!10}66.67 & \cellcolor{yellow!10}54.00 & \cellcolor{yellow!10}47.37 & \cellcolor{yellow!10}46.43 & \cellcolor{yellow!10}53.28 & \cellcolor{yellow!10}55.56 & \cellcolor{orange!10}33.56 & \cellcolor{orange!10}40.44 & \cellcolor{teal!10}50.00 & \cellcolor{teal!10}28.04 & \cellcolor{red!10}76.00 & \cellcolor{red!10}31.03 & \cellcolor{red!10}100.00 & \cellcolor{gray!10}37.56 \\
Gemma3-27B-it \cite{gemmateam2025gemma3technicalreport} & \cellcolor{blue!10}100.00 & \cellcolor{blue!10}50.00 & \cellcolor{green!10}0.00 & \cellcolor{green!10}38.46 & \cellcolor{green!10}83.33 & \cellcolor{yellow!10}48.40 & \cellcolor{yellow!10}31.58 & \cellcolor{yellow!10}25.00 & \cellcolor{yellow!10}41.92 & \cellcolor{yellow!10}40.74 & \cellcolor{orange!10}32.88 & \cellcolor{orange!10}47.79 & \cellcolor{teal!10}50.00 & \cellcolor{teal!10}32.82 & \cellcolor{red!10}54.00 & \cellcolor{red!10}31.03 & \cellcolor{red!10}80.00 & \cellcolor{gray!10}38.02 \\
QVQ-72B-Preview \cite{qvq-72b-preview} & \cellcolor{blue!10}100.00 & \cellcolor{blue!10}43.33 & \cellcolor{green!10}0.00 & \cellcolor{green!10}46.15 & \cellcolor{green!10}83.33 & \cellcolor{yellow!10}58.00 & \cellcolor{yellow!10}42.11 & \cellcolor{yellow!10}46.43 & \cellcolor{yellow!10}44.10 & \cellcolor{yellow!10}62.96 & \cellcolor{orange!10}36.30 & \cellcolor{orange!10}48.16 & \cellcolor{teal!10}50.00 & \cellcolor{teal!10}28.55 & \cellcolor{red!10}78.00 & \cellcolor{red!10}48.28 & \cellcolor{red!10}100.00 & \cellcolor{gray!10}39.13 \\
LLaVA-OneVision-72B \cite{llava-onevisoin} & \cellcolor{blue!10}0.00 & \cellcolor{blue!10}33.33 & \cellcolor{green!10}0.00 & \cellcolor{green!10}26.92 & \cellcolor{green!10}66.67 & \cellcolor{yellow!10}61.20 & \cellcolor{yellow!10}57.89 & \cellcolor{yellow!10}57.14 & \cellcolor{yellow!10}60.70 & \cellcolor{yellow!10}51.85 & \cellcolor{orange!10}41.10 & \cellcolor{orange!10}60.29 & \cellcolor{teal!10}100.00 & \cellcolor{teal!10}38.24 & \cellcolor{red!10}82.00 & \cellcolor{red!10}41.38 & \cellcolor{red!10}80.00 & \cellcolor{gray!10}47.67 \\
Qwen2.5-VL-72B-Instruct \cite{qwen-2.5-vl} & \cellcolor{blue!10}0.00 & \cellcolor{blue!10}40.67 & \cellcolor{green!10}0.00 & \cellcolor{green!10}53.85 & \cellcolor{green!10}50.00 & \cellcolor{yellow!10}68.00 & \cellcolor{yellow!10}68.42 & \cellcolor{yellow!10}53.57 & \cellcolor{yellow!10}55.02 & \cellcolor{yellow!10}74.07 & \cellcolor{orange!10}58.22 & \cellcolor{orange!10}60.66 & \cellcolor{teal!10}60.00 & \cellcolor{teal!10}35.53 & \cellcolor{red!10}76.00 & \cellcolor{red!10}43.10 & \cellcolor{red!10}100.00 & \cellcolor{gray!10}48.08 \\
\bottomrule
\end{tabular}
}
\label{tab:main_results}
\end{table*}

% \begin{table*}[]
% \resizebox{\textwidth}{!}{
% \begin{tabular}{llllllll}
% \toprule                         & Time \& Calendar & Space \& Location & Geometry \& Shapes & Objects \& Motion & Reasoning \& Observation & Organization \& Pattern & Avg. \\ \midrule
% Random Guass                   &    &    &    &    &    &    &      \\
% Human           &    &    &    &    &    &    &      \\
% \multicolumn{8}{c}{\cellcolor{gray!20}\textit{\textbf{Close-Sourced Models}}} \\
% GPT-4o                   &    &    &    &    &    &    &      \\
% Gemini-1.5-Pro           &    &    &    &    &    &    &      \\
% Claude-3.5-Sonnet        &    &    &    &    &    &    &      \\ 
% \multicolumn{8}{c}{\cellcolor{gray!20}\textit{\textbf{Close-Sourced Models}}} \\
% InterVL2.5-8B            &    &    &    &    &    &    &      \\
% InternVL2.5-78B          &    &    &    &    &    &    &      \\
% InternLM-Xcomposer2.5-7B &    &    &    &    &    &    &      \\
% LLaVA-Interleave-7B      &    &    &    &    &    &    &      \\
% LLaVA-OneVision-7B       &    &    &    &    &    &    &      \\
% LLaVA-OneVision-72B      &    &    &    &    &    &    & \\ \bottomrule
% \end{tabular}
% }
% \end{table*}

\section{Experiment}

\subsection{Main Results}
There are a total of 17 subtasks for the evaluation from the perspectives of Temporal Reasoning, Spatial Reasoning, Geometric Reasoning, Logical Reasoning, and Pattern Recognition abilities over 21 VLMs. Table \ref{tab:main_results} provides detailed evaluation results across six visual reasoning tasks. Human performance is near-perfect with an average score of 93.30, while random guessing achieves only 29.83, which emphasizes that these tasks, though inherently solvable by humans, pose substantial challenges to current AI systems.

%In contrast, Claude-3.7-Sonnet (46.63\%), Qwen-VL-Max (47.03\%), and Gemini2.0-Flash (49.77\%) achieve notably higher performance. 
Figure~\ref{fig:radar} shows the comparative performance of six various prominent LVLMs across six tasks. Their relative strengths lie particularly in tasks requiring spatial reasoning and observational interpretation, suggesting these models have better internal representations or more effective cross-modal alignment between visual and linguistic information. However, despite these advancements, even these top-performing closed-source models exhibit notable shortcomings relative to humans, particularly in high-complexity reasoning scenarios (e.g., Geometry and Objects and Motion), reflecting an ongoing gap in advanced spatial reasoning, logical reasoning and pattern recognition capabilities.

Open-source models present an even more heterogeneous and generally lower performance landscape, indicative of diverse model architectures, varying degrees of multi-modal integration sophistication, and potentially inconsistent data quality or quantity during training. For example, large open-source models, including Qwen2.5-VL-72B-Instruct (48.08\%) and LLaVA-OneVision-72B (47.67\%), demonstrate performance comparable to mid-tier closed-source models. Their comparatively stronger results, particularly in Geometry and Shapes and Organization and Pattern tasks, suggest these models benefit from scale and possibly more sophisticated visual encoders or pre-training strategies. However, they still encounter substantial difficulties in tasks requiring nuanced observation or reasoning about motion and object interactions, highlighting remaining challenges in achieving cognative visual reasoning. The variability across different tasks, especially pronounced in Objects and Motion and Reasoning and Observation categories, points toward crucial areas requiring further research: enhancing temporal reasoning, improving dynamic visual understanding, and strengthening the integration of geometric and spatial cognition into visual-language models.

\begin{table*}[t]
\centering
\caption{Performance comparison of vision-language models across different categories in single-image settings. The rightmost column
shows the performance improvement ratio when switching from multi-image to single-image settings.}
\resizebox{\textwidth}{!}{
\begin{tabular}{lccccccccccccccccccc}
\toprule
\multirow{2}{*}{Models} & \multicolumn{2}{c}{Time and Calendar} & \multicolumn{3}{c}{Space and Location} & \multicolumn{5}{c}{Geometry and Shapes} & \multicolumn{2}{c}{Objects and Motion} & \multicolumn{2}{c}{Reasoning and Observation} & \multicolumn{3}{c}{Organization and Pattern} & \multirow{2}{*}{Avg.} & \multirow{2}{*}{\begin{tabular}[c]{@{}c@{}}Improvement\\ Ratio\end{tabular}} \\ 
                        & Calender & Clock & Direction & Location & Place & Angle & Quad & Rectangular & Shape & Triangle & Cube & Move & Reasoning & Observe & Organize & Pattern & Weight & & \\ 
\midrule
Random Guess            & \cellcolor{blue!10}33.33 & \cellcolor{blue!10}32.78 & \cellcolor{green!10}25.00 & \cellcolor{green!10}29.81 & \cellcolor{green!10}33.33 & \cellcolor{yellow!10}31.00 & \cellcolor{yellow!10}27.63 & \cellcolor{yellow!10}29.17 & \cellcolor{yellow!10}31.84 & \cellcolor{yellow!10}29.01 & \cellcolor{orange!10}28.37 & \cellcolor{orange!10}29.35 & \cellcolor{teal!10}33.33 & \cellcolor{teal!10}29.41 & \cellcolor{red!10}30.17 & \cellcolor{red!10}31.32 & \cellcolor{red!10}33.33 & \cellcolor{gray!10}29.83 & \cellcolor{gray!10}- \\ 
Human                   & \cellcolor{blue!10}100.00 & \cellcolor{blue!10}96.00 & \cellcolor{green!10}100.00 & \cellcolor{green!10}93.85 & \cellcolor{green!10}96.67 & \cellcolor{yellow!10}95.60 & \cellcolor{yellow!10}96.84 & \cellcolor{yellow!10}95.00 & \cellcolor{yellow!10}94.02 & \cellcolor{yellow!10}94.07 & \cellcolor{orange!10}97.67 & \cellcolor{orange!10}94.63 & \cellcolor{teal!10}100.00 & \cellcolor{teal!10}93.59 & \cellcolor{red!10}93.20 & \cellcolor{red!10}95.52 & \cellcolor{red!10}100.00 & \cellcolor{gray!10}93.30 & \cellcolor{gray!10}- \\ 
\multicolumn{20}{c}{\cellcolor{gray!20}\textit{\textbf{Close-Source Models}}} \\ 
GPT-4o-mini \cite{openai2024gpt4ocard} & \cellcolor{blue!10}100.00 & \cellcolor{blue!10}20.00 & \cellcolor{green!10}0.00 & \cellcolor{green!10}30.77 & \cellcolor{green!10}100.00 & \cellcolor{yellow!10}42.80 & \cellcolor{yellow!10}26.32 & \cellcolor{yellow!10}50.00 & \cellcolor{yellow!10}56.77 & \cellcolor{yellow!10}40.74 & \cellcolor{orange!10}34.93 & \cellcolor{orange!10}43.01 & \cellcolor{teal!10}90.00 & \cellcolor{teal!10}32.43 & \cellcolor{red!10}72.00 & \cellcolor{red!10}37.93 & \cellcolor{red!10}60.00 & \cellcolor{gray!10}39.65 & \cellcolor{gray!10}\textcolor{ForestGreen}{13.7\%} \\ 
GPT-4o \cite{openai2024gpt4ocard} & \cellcolor{blue!10}80.00 & \cellcolor{blue!10}40.67 & \cellcolor{green!10}100.00 & \cellcolor{green!10}42.31 & \cellcolor{green!10}66.67 & \cellcolor{yellow!10}68.40 & \cellcolor{yellow!10}57.89 & \cellcolor{yellow!10}64.29 & \cellcolor{yellow!10}68.12 & \cellcolor{yellow!10}44.44 & \cellcolor{orange!10}42.47 & \cellcolor{orange!10}56.99 & \cellcolor{teal!10}60.00 & \cellcolor{teal!10}30.10 & \cellcolor{red!10}90.40 & \cellcolor{red!10}44.83 & \cellcolor{red!10}100.00 & \cellcolor{gray!10}45.52 & \cellcolor{gray!10}\textcolor{ForestGreen}{12.9\%} \\ 
Claude-3.7-Sonnet \cite{calude} & \cellcolor{blue!10}100.00 & \cellcolor{blue!10}54.67 & \cellcolor{green!10}80.00 & \cellcolor{green!10}65.38 & \cellcolor{green!10}83.33 & \cellcolor{yellow!10}61.20 & \cellcolor{yellow!10}68.42 & \cellcolor{yellow!10}78.57 & \cellcolor{yellow!10}68.56 & \cellcolor{yellow!10}77.78 & \cellcolor{orange!10}43.84 & \cellcolor{orange!10}69.12 & \cellcolor{teal!10}100.00 & \cellcolor{teal!10}34.37 & \cellcolor{red!10}92.00 & \cellcolor{red!10}63.79 & \cellcolor{red!10}100.00 & \cellcolor{gray!10}51.69 & \cellcolor{gray!10}\textcolor{ForestGreen}{10.8\%} \\ 
Gemini2.0-Flash \cite{gemini-2.0} & \cellcolor{blue!10}20.00 & \cellcolor{blue!10}76.67 & \cellcolor{green!10}100.00 & \cellcolor{green!10}61.54 & \cellcolor{green!10}83.33 & \cellcolor{yellow!10}58.00 & \cellcolor{yellow!10}63.16 & \cellcolor{yellow!10}42.86 & \cellcolor{yellow!10}71.62 & \cellcolor{yellow!10}59.26 & \cellcolor{orange!10}46.58 & \cellcolor{orange!10}73.90 & \cellcolor{teal!10}100.00 & \cellcolor{teal!10}39.41 & \cellcolor{red!10}90.00 & \cellcolor{red!10}46.55 & \cellcolor{red!10}100.00 & \cellcolor{gray!10}53.90 & \cellcolor{gray!10}\textcolor{ForestGreen}{8.3\%} \\ 
Qwen-VL-Max \cite{qwen-vl-max} & \cellcolor{blue!10}0.00 & \cellcolor{blue!10}53.33 & \cellcolor{green!10}100.00 & \cellcolor{green!10}73.08 & \cellcolor{green!10}83.33 & \cellcolor{yellow!10}80.00 & \cellcolor{yellow!10}52.63 & \cellcolor{yellow!10}75.00 & \cellcolor{yellow!10}69.87 & \cellcolor{yellow!10}66.67 & \cellcolor{orange!10}57.53 & \cellcolor{orange!10}72.43 & \cellcolor{teal!10}100.00 & \cellcolor{teal!10}43.54 & \cellcolor{red!10}91.60 & \cellcolor{red!10}41.38 & \cellcolor{red!10}80.00 & \cellcolor{gray!10}57.03 & \cellcolor{gray!10}\textcolor{ForestGreen}{21.3\%} \\ 
\multicolumn{20}{c}{\cellcolor{gray!20}\textit{\textbf{Open-Source Models}}} \\ 
Idefics3-8B \cite{idefics} & \cellcolor{blue!10}0.00 & \cellcolor{blue!10}10.00 & \cellcolor{green!10}20.00 & \cellcolor{green!10}11.54 & \cellcolor{green!10}16.67 & \cellcolor{yellow!10}10.00 & \cellcolor{yellow!10}5.26 & \cellcolor{yellow!10}32.14 & \cellcolor{yellow!10}20.52 & \cellcolor{yellow!10}7.41 & \cellcolor{orange!10}17.12 & \cellcolor{orange!10}18.01 & \cellcolor{teal!10}0.00 & \cellcolor{teal!10}12.53 & \cellcolor{red!10}30.00 & \cellcolor{red!10}20.69 & \cellcolor{red!10}0.00 & \cellcolor{gray!10}15.64 & \cellcolor{gray!10}\textcolor{ForestGreen}{21.2\%} \\ 
LLaMA-3.2-90B-Vision-Instruct \cite{meta2024llama32} & \cellcolor{blue!10}80.00 & \cellcolor{blue!10}30.00 & \cellcolor{green!10}0.00 & \cellcolor{green!10}15.38 & \cellcolor{green!10}33.33 & \cellcolor{yellow!10}26.00 & \cellcolor{yellow!10}15.79 & \cellcolor{yellow!10}25.00 & \cellcolor{yellow!10}17.03 & \cellcolor{yellow!10}33.33 & \cellcolor{orange!10}27.40 & \cellcolor{orange!10}26.47 & \cellcolor{teal!10}100.00 & \cellcolor{teal!10}19.64 & \cellcolor{red!10}49.60 & \cellcolor{red!10}12.07 & \cellcolor{red!10}0.00 & \cellcolor{gray!10}22.38 & \cellcolor{gray!10}\textcolor{red}{-11.9\%} \\ 
Emu2-Chat \cite{emu2} & \cellcolor{blue!10}60.00 & \cellcolor{blue!10}12.67 & \cellcolor{green!10}100.00 & \cellcolor{green!10}23.08 & \cellcolor{green!10}16.67 & \cellcolor{yellow!10}24.00 & \cellcolor{yellow!10}42.11 & \cellcolor{yellow!10}28.57 & \cellcolor{yellow!10}24.02 & \cellcolor{yellow!10}18.52 & \cellcolor{orange!10}22.60 & \cellcolor{orange!10}24.63 & \cellcolor{teal!10}10.00 & \cellcolor{teal!10}22.87 & \cellcolor{red!10}12.00 & \cellcolor{red!10}22.41 & \cellcolor{red!10}0.00 & \cellcolor{gray!10}23.08 & \cellcolor{gray!10}\textcolor{ForestGreen}{281.5\%} \\ 
DeepSeek-VL2 \cite{deepseek-vl-2} & \cellcolor{blue!10}20.00 & \cellcolor{blue!10}33.33 & \cellcolor{green!10}0.00 & \cellcolor{green!10}19.23 & \cellcolor{green!10}33.33 & \cellcolor{yellow!10}28.00 & \cellcolor{yellow!10}10.53 & \cellcolor{yellow!10}32.14 & \cellcolor{yellow!10}32.31 & \cellcolor{yellow!10}25.93 & \cellcolor{orange!10}13.70 & \cellcolor{orange!10}32.35 & \cellcolor{teal!10}0.00 & \cellcolor{teal!10}20.03 & \cellcolor{red!10}46.00 & \cellcolor{red!10}27.59 & \cellcolor{red!10}100.00 & \cellcolor{gray!10}24.77 & \cellcolor{gray!10}\textcolor{ForestGreen}{60.1\%} \\ 
Mantis-CLIP \cite{mantis} & \cellcolor{blue!10}0.00 & \cellcolor{blue!10}35.33 & \cellcolor{green!10}80.00 & \cellcolor{green!10}23.08 & \cellcolor{green!10}0.00 & \cellcolor{yellow!10}28.00 & \cellcolor{yellow!10}42.11 & \cellcolor{yellow!10}46.43 & \cellcolor{yellow!10}31.88 & \cellcolor{yellow!10}11.11 & \cellcolor{orange!10}26.03 & \cellcolor{orange!10}25.00 & \cellcolor{teal!10}0.00 & \cellcolor{teal!10}27.52 & \cellcolor{red!10}12.00 & \cellcolor{red!10}34.48 & \cellcolor{red!10}0.00 & \cellcolor{gray!10}27.50 & \cellcolor{gray!10}\textcolor{red}{-9.0\%} \\ 
LLaVA-Interleave-7B\cite{li2024llavanextinterleavetacklingmultiimagevideo} & \cellcolor{blue!10}00.00 & \cellcolor{blue!10}30.00 & \cellcolor{green!10}100.00 & \cellcolor{green!10}30.77 & \cellcolor{green!10}0.00 & \cellcolor{yellow!10}26.00 & \cellcolor{yellow!10}36.84 & \cellcolor{yellow!10}42.86 & \cellcolor{yellow!10}33.19 & \cellcolor{yellow!10}14.81 & \cellcolor{orange!10}31.51 & \cellcolor{orange!10}26.47 & \cellcolor{teal!10}50.00 & \cellcolor{teal!10}29.07 & \cellcolor{red!10}28.00 & \cellcolor{red!10}25.86 & \cellcolor{red!10}0.00 & \cellcolor{gray!10}29.24 & \cellcolor{gray!10}\textcolor{red}{-17.6\%} \\ 
Phi-3.5-vision-instruct \cite{phi-3.5} & \cellcolor{blue!10}0.00 & \cellcolor{blue!10}13.33 & \cellcolor{green!10}80.00 & \cellcolor{green!10}19.23 & \cellcolor{green!10}16.67 & \cellcolor{yellow!10}24.40 & \cellcolor{yellow!10}10.53 & \cellcolor{yellow!10}42.86 & \cellcolor{yellow!10}34.50 & \cellcolor{yellow!10}22.22 & \cellcolor{orange!10}32.19 & \cellcolor{orange!10}29.78 & \cellcolor{teal!10}20.00 & \cellcolor{teal!10}31.40 & \cellcolor{red!10}46.00 & \cellcolor{red!10}25.86 & \cellcolor{red!10}100.00 & \cellcolor{gray!10}30.93 & \cellcolor{gray!10}\textcolor{ForestGreen}{36.1\%} \\ 
LLaVA-OneVision-7B \cite{llava-onevisoin} & \cellcolor{blue!10}0.00 & \cellcolor{blue!10}43.33 & \cellcolor{green!10}0.00 & \cellcolor{green!10}23.08 & \cellcolor{green!10}100.00 & \cellcolor{yellow!10}44.00 & \cellcolor{yellow!10}21.05 & \cellcolor{yellow!10}35.71 & \cellcolor{yellow!10}44.10 & \cellcolor{yellow!10}44.44 & \cellcolor{orange!10}30.82 & \cellcolor{orange!10}42.65 & \cellcolor{teal!10}40.00 & \cellcolor{teal!10}29.07 & \cellcolor{red!10}64.40 & \cellcolor{red!10}27.59 & \cellcolor{red!10}80.00 & \cellcolor{gray!10}35.47 & \cellcolor{gray!10}\textcolor{red}{-3.2\%} \\ 
InternVL2.5-8B \cite{internvl-2.5} & \cellcolor{blue!10}0.00 & \cellcolor{blue!10}33.33 & \cellcolor{green!10}0.00 & \cellcolor{green!10}26.92 & \cellcolor{green!10}50.00 & \cellcolor{yellow!10}46.40 & \cellcolor{yellow!10}31.58 & \cellcolor{yellow!10}39.29 & \cellcolor{yellow!10}51.53 & \cellcolor{yellow!10}48.15 & \cellcolor{orange!10}31.51 & \cellcolor{orange!10}42.65 & \cellcolor{teal!10}30.00 & \cellcolor{teal!10}28.42 & \cellcolor{red!10}60.80 & \cellcolor{red!10}29.31 & \cellcolor{red!10}80.00 & \cellcolor{gray!10}36.16 & \cellcolor{gray!10}\textcolor{ForestGreen}{46.3\%} \\ 
Gemma3-27B-it\cite{gemmateam2025gemma3technicalreport} & \cellcolor{blue!10}80.00 & \cellcolor{blue!10}40.00 & \cellcolor{green!10}0.00 & \cellcolor{green!10}26.92 & \cellcolor{green!10}33.33 & \cellcolor{yellow!10}48.40 & \cellcolor{yellow!10}21.05 & \cellcolor{yellow!10}57.14 & \cellcolor{yellow!10}45.85 & \cellcolor{yellow!10}33.33 & \cellcolor{orange!10}33.56 & \cellcolor{orange!10}45.22 & \cellcolor{teal!10}100.00 & \cellcolor{teal!10}30.10 & \cellcolor{red!10}66.80 & \cellcolor{red!10}20.69 & \cellcolor{red!10}60.00 & \cellcolor{gray!10}36.80 & \cellcolor{gray!10}\textcolor{ForestGreen}{2.1\%} \\ 
Kimi-VL-A3B-Thinking\cite{kimiteam2025kimivltechnicalreport} & \cellcolor{blue!10}0.00 & \cellcolor{blue!10}33.33 & \cellcolor{green!10}0.00 & \cellcolor{green!10}34.62 & \cellcolor{green!10}50.00 & \cellcolor{yellow!10}62.00 & \cellcolor{yellow!10}52.63 & \cellcolor{yellow!10}39.29 & \cellcolor{yellow!10}52.40 & \cellcolor{yellow!10}77.78 & \cellcolor{orange!10}26.03 & \cellcolor{orange!10}55.15 & \cellcolor{teal!10}50.00 & \cellcolor{teal!10}25.19 & \cellcolor{red!10}86.00 & \cellcolor{red!10}39.66 & \cellcolor{red!10}100.00 & \cellcolor{gray!10}38.72 & \cellcolor{gray!10}\textcolor{ForestGreen}{13.4\%} \\
LLaVA-OneVision-72B \cite{llava-onevisoin} & \cellcolor{blue!10}20.00 & \cellcolor{blue!10}53.33 & \cellcolor{green!10}0.00 & \cellcolor{green!10}30.77 & \cellcolor{green!10}33.33 & \cellcolor{yellow!10}38.00 & \cellcolor{yellow!10}47.37 & \cellcolor{yellow!10}39.29 & \cellcolor{yellow!10}51.53 & \cellcolor{yellow!10}55.56 & \cellcolor{orange!10}39.73 & \cellcolor{orange!10}41.54 & \cellcolor{teal!10}100.00 & \cellcolor{teal!10}32.95 & \cellcolor{red!10}32.00 & \cellcolor{red!10}55.17 & \cellcolor{red!10}100.00 & \cellcolor{gray!10}39.24 & \cellcolor{gray!10}\textcolor{red}{-17.7\%} \\ 
Mistral-Small-3.1-24B-Instruct \cite{mistral} & \cellcolor{blue!10}20.00 & \cellcolor{blue!10}40.00 & \cellcolor{green!10}0.00 & \cellcolor{green!10}38.46 & \cellcolor{green!10}50.00 & \cellcolor{yellow!10}64.00 & \cellcolor{yellow!10}57.89 & \cellcolor{yellow!10}46.43 & \cellcolor{yellow!10}56.77 & \cellcolor{yellow!10}70.37 & \cellcolor{orange!10}30.14 & \cellcolor{orange!10}50.74 & \cellcolor{teal!10}100.00 & \cellcolor{teal!10}31.65 & \cellcolor{red!10}82.00 & \cellcolor{red!10}43.10 & \cellcolor{red!10}80.00 & \cellcolor{gray!10}42.21 & \cellcolor{gray!10}\textcolor{ForestGreen}{34.7\%} \\ 
QVQ-72B-Preview \cite{qvq-72b-preview} & \cellcolor{blue!10}80.00 & \cellcolor{blue!10}41.33 & \cellcolor{green!10}80.00 & \cellcolor{green!10}61.54 & \cellcolor{green!10}50.00 & \cellcolor{yellow!10}64.00 & \cellcolor{yellow!10}68.42 & \cellcolor{yellow!10}39.29 & \cellcolor{yellow!10}58.95 & \cellcolor{yellow!10}81.48 & \cellcolor{orange!10}32.19 & \cellcolor{orange!10}64.34 & \cellcolor{teal!10}50.00 & \cellcolor{teal!10}35.01 & \cellcolor{red!10}90.00 & \cellcolor{red!10}50.00 & \cellcolor{red!10}100.00 & \cellcolor{gray!10}47.44 & \cellcolor{gray!10}\textcolor{ForestGreen}{21.2\%} \\ 
InternVL2.5-78B \cite{internvl-2.5} & \cellcolor{blue!10}80.00 & \cellcolor{blue!10}50.00 & \cellcolor{green!10}100.00 & \cellcolor{green!10}42.31 & \cellcolor{green!10}50.00 & \cellcolor{yellow!10}62.80 & \cellcolor{yellow!10}63.16 & \cellcolor{yellow!10}57.14 & \cellcolor{yellow!10}65.94 & \cellcolor{yellow!10}55.56 & \cellcolor{orange!10}32.19 & \cellcolor{orange!10}61.76 & \cellcolor{teal!10}90.00 & \cellcolor{teal!10}36.43 & \cellcolor{red!10}88.00 & \cellcolor{red!10}36.21 & \cellcolor{red!10}100.00 & \cellcolor{gray!10}47.73 & \cellcolor{gray!10}\textcolor{ForestGreen}{27.1\%} \\
Kimi-VL-A3B-Instruct\cite{kimiteam2025kimivltechnicalreport} & \cellcolor{blue!10}0.00 & \cellcolor{blue!10}70.00 & \cellcolor{green!10}100.00 & \cellcolor{green!10}50.00 & \cellcolor{green!10}66.67 & \cellcolor{yellow!10}50.00 & \cellcolor{yellow!10}31.58 & \cellcolor{yellow!10}35.71 & \cellcolor{yellow!10}59.39 & \cellcolor{yellow!10}51.85 & \cellcolor{orange!10}46.58 & \cellcolor{orange!10}62.13 & \cellcolor{teal!10}50.00 & \cellcolor{teal!10}38.11 & \cellcolor{red!10}82.00 & \cellcolor{red!10}46.55 & \cellcolor{red!10}100.00 & \cellcolor{gray!10}48.37 &\cellcolor{gray!10}\textcolor{ForestGreen}{29.6\%} \\
Qwen2.5-VL-7B-Instruct \cite{qwen-2.5-vl} & \cellcolor{blue!10}0.00 & \cellcolor{blue!10}53.33 & \cellcolor{green!10}100.00 & \cellcolor{green!10}46.15 & \cellcolor{green!10}83.33 & \cellcolor{yellow!10}72.80 & \cellcolor{yellow!10}52.63 & \cellcolor{yellow!10}60.71 & \cellcolor{yellow!10}61.14 & \cellcolor{yellow!10}55.56 & \cellcolor{orange!10}60.96 & \cellcolor{orange!10}64.34 & \cellcolor{teal!10}100.00 & \cellcolor{teal!10}37.86 & \cellcolor{red!10}92.00 & \cellcolor{red!10}36.21 & \cellcolor{red!10}80.00 & \cellcolor{gray!10}51.10 & \cellcolor{gray!10}\textcolor{ForestGreen}{74.8\%} \\ 
Qwen2.5-VL-72B-Instruct \cite{qwen-2.5-vl} & \cellcolor{blue!10}20.00 & \cellcolor{blue!10}55.33 & \cellcolor{green!10}100.00 & \cellcolor{green!10}73.08 & \cellcolor{green!10}83.33 & \cellcolor{yellow!10}80.00 & \cellcolor{yellow!10}52.63 & \cellcolor{yellow!10}75.00 & \cellcolor{yellow!10}69.87 & \cellcolor{yellow!10}66.67 & \cellcolor{orange!10}57.53 & \cellcolor{orange!10}72.43 & \cellcolor{teal!10}90.00 & \cellcolor{teal!10}43.54 & \cellcolor{red!10}92.00 & \cellcolor{red!10}41.38 & \cellcolor{red!10}100.00 & \cellcolor{gray!10}57.03 & \cellcolor{gray!10}\textcolor{ForestGreen}{18.6\%} \\ 
\multicolumn{20}{c}{\cellcolor{gray!20}\textit{\textbf{Math Specialist Models}}} \\ 
G-LLaVA-13B \cite{g-llava} & \cellcolor{blue!10}0.00 & \cellcolor{blue!10}40.00 & \cellcolor{green!10}0.00 & \cellcolor{green!10}23.08 & \cellcolor{green!10}33.33 & \cellcolor{yellow!10}20.40 & \cellcolor{yellow!10}31.58 & \cellcolor{yellow!10}32.14 & \cellcolor{yellow!10}26.64 & \cellcolor{yellow!10}25.93 & \cellcolor{orange!10}15.75 & \cellcolor{orange!10}26.10 & \cellcolor{teal!10}0.00 & \cellcolor{teal!10}26.49 & \cellcolor{red!10}24.00 & \cellcolor{red!10}24.14 & \cellcolor{red!10}20.00 & \cellcolor{gray!10}25.47 & \cellcolor{gray!10}-\\ 
G-LLaVA-7B \cite{g-llava} & \cellcolor{blue!10}100.00 & \cellcolor{blue!10}36.67 & \cellcolor{green!10}20.00 & \cellcolor{green!10}30.77 & \cellcolor{green!10}0.00 & \cellcolor{yellow!10}30.00 & \cellcolor{yellow!10}21.05 & \cellcolor{yellow!10}50.00 & \cellcolor{yellow!10}31.88 & \cellcolor{yellow!10}40.74 & \cellcolor{orange!10}23.97 & \cellcolor{orange!10}27.21 & \cellcolor{teal!10}0.00 & \cellcolor{teal!10}27.26 & \cellcolor{red!10}28.00 & \cellcolor{red!10}24.14 & \cellcolor{red!10}100.00 & \cellcolor{gray!10}28.26 & \cellcolor{gray!10}-\\ 
MathLlava \cite{math-llava} & \cellcolor{blue!10}100.00 & \cellcolor{blue!10}20.00 & \cellcolor{green!10}80.00 & \cellcolor{green!10}26.92 & \cellcolor{green!10}0.00 & \cellcolor{yellow!10}32.00 & \cellcolor{yellow!10}31.58 & \cellcolor{yellow!10}21.43 & \cellcolor{yellow!10}27.51 & \cellcolor{yellow!10}11.11 & \cellcolor{orange!10}34.93 & \cellcolor{orange!10}29.04 & \cellcolor{teal!10}40.00 & \cellcolor{teal!10}29.97 & \cellcolor{red!10}28.40 & \cellcolor{red!10}29.31 & \cellcolor{red!10}80.00 & \cellcolor{gray!10}29.30 & \cellcolor{gray!10}-\\ 
\bottomrule
\end{tabular}
}
\label{tab:results_single}
\end{table*}

\begin{table*}[t]
\centering
\caption{Influence of Chain-of-Thought \cite{cot} on model performances.}
\resizebox{\textwidth}{!}{
\begin{tabular}{l c  cc  ccc  ccccc  cc  cc  ccc  c}
\toprule
\multirow{2}{*}{\textbf{Model}} & \multirow{2}{*}{\textbf{CoT}} & \multicolumn{2}{c}{\textbf{Time and Calendar}} & \multicolumn{3}{c}{\textbf{Space and Location}} & \multicolumn{5}{c}{\textbf{Geometry and Shapes}} & \multicolumn{2}{c}{\textbf{Objects and Motion}} & \multicolumn{2}{c}{\textbf{Reasoning and Observation}} & \multicolumn{3}{c}{\textbf{Organization and Pattern}} & \multirow{2}{*}{\textbf{Avg.}} \\
 &  & \textbf{Calender} & \textbf{Clock} & \textbf{Direction} & \textbf{Location} & \textbf{Place} & \textbf{Angle} & \textbf{Quad} & \textbf{Rectangular} & \textbf{Shape} & \textbf{Triangle} & \textbf{Cube} & \textbf{Move} & \textbf{Reasoning} & \textbf{Observe} & \textbf{Organize} & \textbf{Pattern} & \textbf{Weight} & \\
\midrule
\multirow{3}{*}{GPT-4o \cite{openai2024gpt4ocard}} 
    & \ding{55} & 100.00 & 40.00 & 20.00  & 30.77 & 66.67 & 46.00 & 57.89 & 28.57 & 50.22 & 51.85 & 37.67 & 50.37 & 90.00 & 31.27 & 76.00 & 37.93 & 80.00 & 40.29 \\
    & \ding{51} & 100.00&40.00&0.00&38.46&66.67&52.00&63.16&32.14&53.71&66.67&33.56&52.57&100.00&30.75&82.00&58.62&100.00
& 42.03 \\
     & & \textbf{0.00} & \textbf{0.00} & \textcolor{red}{\textbf{-20.00}} & \textcolor{ForestGreen}{\textbf{+7.69}} & \textbf{0.00} & \textcolor{ForestGreen}{\textbf{+6.00}} & \textcolor{ForestGreen}{\textbf{+5.27}} & \textcolor{ForestGreen}{\textbf{+3.57}} & \textcolor{ForestGreen}{\textbf{+3.49}} & \textcolor{ForestGreen}{\textbf{+14.82}} & \textcolor{red}{\textbf{-4.11}} & \textcolor{ForestGreen}{\textbf{+2.20}} & \textcolor{ForestGreen}{\textbf{+10.00}} & \textcolor{red}{\textbf{-0.52}} & \textcolor{ForestGreen}{\textbf{+6.00}} & \textcolor{ForestGreen}{\textbf{+20.69}} & \textcolor{ForestGreen}{\textbf{+20.00}} & \textcolor{ForestGreen}{\textbf{+1.74}} \\
\midrule
\multirow{3}{*}{Qwen-VL-Max \cite{qwen-vl-max}} 
    & \ding{55} & 0.00   & 46.67 & 0.00  & 42.31 & 66.67 & 74.00 & 52.63 & 42.86 & 54.15 & 66.67 & 56.16 & 60.66 & 50.00  & 35.27 & 68.00 & 39.66 & 100.00 & 47.03 \\
    & \ding{51} &20.00&36.67&100.00&57.69&66.67&74.40&52.63&57.14&60.26&77.78&52.74&61.03&90.00&36.05&93.60&44.83&100.00
& 49.48 \\
    &  & \textcolor{ForestGreen}{\textbf{+20.00}} & \textcolor{red}{\textbf{-10.00}} & \textcolor{ForestGreen}{\textbf{+100.00}} & \textcolor{ForestGreen}{\textbf{+15.38}} & \textbf{0.00} & \textcolor{ForestGreen}{\textbf{+0.40}} & \textbf{0.00} & \textcolor{ForestGreen}{\textbf{+14.28}} & \textcolor{ForestGreen}{\textbf{+6.11}} & \textcolor{ForestGreen}{\textbf{+11.11}} & \textcolor{red}{\textbf{-3.42}} & \textcolor{ForestGreen}{\textbf{+0.37}} & \textcolor{ForestGreen}{\textbf{+40.00}} & \textcolor{ForestGreen}{\textbf{+0.78}} & \textcolor{ForestGreen}{\textbf{+25.60}} & \textcolor{ForestGreen}{\textbf{+5.17}} & \textbf{0.00} & \textcolor{ForestGreen}{\textbf{+2.45}} \\
\midrule
\multirow{3}{*}{Gemini2.0-Flash \cite{gemini-2.0}} 
    & \ding{55} & 100.00 & 70.00 & 20.00  & 57.69 & 66.67 & 70.00 & 68.42 & 53.57 & 61.14 & 70.37 & 44.52 & 68.75 & 40.00  & 35.53 & 74.00 & 46.55 & 100.00 & 49.77 \\
    & \ding{51} & 80.00&83.33&20.00&69.23&83.33&66.40&68.42&67.86&71.62&66.67&41.10&70.96&100.00&37.86&89.40&56.90&100.00
& 53.66 \\
    &  & \textcolor{red}{\textbf{-20.00}} & \textcolor{ForestGreen}{\textbf{+13.33}} & \textbf{0.00} & \textcolor{ForestGreen}{\textbf{+11.54}} & \textcolor{ForestGreen}{\textbf{+16.66}} & \textcolor{red}{\textbf{-3.60}} & \textbf{0.00} & \textcolor{ForestGreen}{\textbf{+14.29}} & \textcolor{ForestGreen}{\textbf{+10.48}} & \textcolor{red}{\textbf{-3.70}} & \textcolor{red}{\textbf{-3.42}} & \textcolor{ForestGreen}{\textbf{+2.21}} & \textcolor{ForestGreen}{\textbf{+60.00}} & \textcolor{ForestGreen}{\textbf{+2.33}} & \textcolor{ForestGreen}{\textbf{+15.40}} & \textcolor{ForestGreen}{\textbf{+10.35}} & \textbf{0.00} & \textcolor{ForestGreen}{\textbf{+3.89}} \\

\bottomrule
\end{tabular}
}
\label{tab:results_cot}
\end{table*}

\subsection{Evaluation in Single-Image Setting}
The evaluation is also conducted in a single-image setting for comparison. In single-image setting, we integrate visual and textual elements into a cohesive layout as shown in Figure~\ref{fig:overview}. If a model performs well in single-image but poorly in multi-image, it suggests the model lacks compositional reasoning ability to link separate inputs.
The results in Table~\ref{tab:results_single} reveal two key findings: First, most models perform significantly better in single-image settings compared to multi-image scenarios~(average improvement of +42.3\%), indicating a strong bias toward single-image optimization. For instance, Qwen-VL-Max shows a +21.3\% gain in single-image performance, while models like Emu2-Chat exhibit dramatic improvements~(+281.5\%). Second, specialized multi-image models like LLaVA-Interleave-7B show the opposite trend~(-17.6\% in single-image mode), achieving higher accuracy in multi-image tasks than in single-image ones. This contrast suggests that unlike dedicated multi-image architectures, conventional models struggle to integrate visual information across multiple inputs, highlighting a critical limitation in current vision-language systems. Addressing this gap by effectively leverage cross-image cues for reasoning remains an essential challenge for future research.

\subsection{Results of Math Specialist Models}
The Math Specialist models, which include G-LLaVA-13B, G-LLaVA-7B, and MathLlava, exhibit relatively low overall performance, with average scores ranging from 25.47 to 29.30. Notably, G-LLaVA-13B records the lowest score at 25.47, while MathLlava achieves a slightly higher average of 29.30. Although these models are designed with a focus on mathematical reasoning, their performance across diverse tasks—such as time and calendar, spatial reasoning, and geometric challenges—remains inconsistent. For instance, while G-LLaVA-7B reaches a perfect score (100.00) on the Calendar sub-task, its scores in other categories, such as Clock and certain geometry-related tasks, are considerably lower. 

Furthermore, the results indicate that these Math Specialist models struggle to match the performance of their general-purpose counterparts. Despite showing some strengths—for example, MathLlava scoring 34.93 on the Cube task—these models fall short on several key aspects, including Clock, Location, and reasoning tasks. This pattern underscores the challenge of integrating specialized mathematical capabilities with the broader spectrum of visual understanding.

\begin{table}[]
\centering
\caption{Accuracy comparison of various models on questions categorized by difficulty along with their average performance.}
\resizebox{0.5\textwidth}{!}{
\begin{tabular}{lllll}
\toprule
\textbf{Models} & \textbf{Easy} & \textbf{Medium} & \textbf{Hard} & \textbf{Avg.} \\ \midrule
LLaMA-3.2-90B-Vision-Instruct               & 22.22 & 26.15 & 23.89 & 25.41 \\ 
Mantis-CLIP               & 29.63 & 29.30 & 32.37 & 30.23 \\
InternVL2.5-78B               & 25.93 & 36.03 & 41.62 & 37.56 \\
QVQ-72B-Preview                & 18.52 & 36.71 & 45.66 & 39.13 \\
LLaVA-OneVision-72B               & 29.63 & 45.32 & 53.76 & 47.62 \\
Qwen2.5-VL-72B-Instruct               & 25.93 & 45.49 & 55.11 & 48.08 \\
\bottomrule
\end{tabular}
}
\label{tab:difficulty_comparison}
\end{table}

\begin{table*}[t]  
    \centering  
    \caption{Comparisons between existing visual math benchmarks for LVLMs.}  
    \label{tab:benchmark_compare}  
    \resizebox{\textwidth}{!}{  
    \begin{tabular}{lcccccccccc}  
        \toprule  
        \multirow{2}{*}{\textbf{Benchmark}} & \multirow{2}{*}{\textbf{Image Numbers}} & \multirow{2}{*}{\textbf{Question Numbers}} & \multicolumn{5}{c}{\textbf{Required Skills}} & \multirow{2}{*}{\textbf{Multi-Images}} & \multirow{2}{*}{\textbf{Answer Type}} \\  
        \cmidrule(r){4-8}  
        & & & \texttt{\textbf{Temporal}} & \texttt{\textbf{Spatial}} & \texttt{\textbf{Geometric}} & \texttt{\textbf{Logical}} & \texttt{\textbf{Pattern}} & & \\  
        \midrule  
        Olympiadbench \cite{he2024olympiadbench} & 5,129 & 8,952 & \textcolor{red}{\ding{55}} & \textcolor{red}{\ding{55}} & \textcolor{ForestGreen}{\ding{51}} & \textcolor{red}{\ding{55}} & \textcolor{red}{\ding{55}} & \textcolor{red}{\ding{55}} & Free-form \\  
        GeoQA \cite{geoqa} & 4,998 & 4,998 & \textcolor{red}{\ding{55}} & \textcolor{red}{\ding{55}} & \textcolor{ForestGreen}{\ding{51}} & \textcolor{red}{\ding{55}} & \textcolor{red}{\ding{55}} & \textcolor{red}{\ding{55}} & Multiple Choice \\  
        MATH-Vision \cite{math_vision} & 3,472 & 3,040 & \textcolor{red}{\ding{55}} & \textcolor{red}{\ding{55}} & \textcolor{ForestGreen}{\ding{51}} & \textcolor{red}{\ding{55}} & \textcolor{red}{\ding{55}} & \textcolor{ForestGreen}{\ding{51}} & Free-form \& Multiple Choice \\  
        MathVista \cite{lu2023mathvista} & 5,487 & 6,141 & \textcolor{ForestGreen}{\ding{51}} & \textcolor{red}{\ding{55}} & \textcolor{ForestGreen}{\ding{51}} & \textcolor{ForestGreen}{\ding{51}} & \textcolor{red}{\ding{55}} & \textcolor{red}{\ding{55}} & Free-form \& Multiple Choice \\  
        $\text{MMMU}_{\text{math}}$ \cite{mmmu} & 577 & 540 & \textcolor{red}{\ding{55}} & \textcolor{red}{\ding{55}} & \textcolor{ForestGreen}{\ding{51}} & \textcolor{red}{\ding{55}} & \textcolor{red}{\ding{55}} & \textcolor{ForestGreen}{\ding{51}} & Free-form \& Multiple Choice \\  
        GeoMath \cite{geomath} & 4,540 & 9,155 & \textcolor{red}{\ding{55}} & \textcolor{red}{\ding{55}} & \textcolor{ForestGreen}{\ding{51}} & \textcolor{red}{\ding{55}} & \textcolor{red}{\ding{55}} & \textcolor{red}{\ding{55}} & Free-form \& Multiple Choice \& Prove \\  
        U-Math \cite{u_math} & 225 & 1,100 & \textcolor{red}{\ding{55}} & \textcolor{red}{\ding{55}} & \textcolor{ForestGreen}{\ding{51}} & \textcolor{red}{\ding{55}} & \textcolor{red}{\ding{55}} & \textcolor{red}{\ding{55}} & Free-form \\  
        Blink \cite{blink} & 7,358 & 3,807 & \textcolor{red}{\ding{55}} & \textcolor{red}{\ding{55}} & \textcolor{red}{\ding{55}} & \textcolor{red}{\ding{55}} & \textcolor{ForestGreen}{\ding{51}} & \textcolor{ForestGreen}{\ding{51}} & Multiple Choice \\  
        MM-MATH \cite{mm_math} & 5,929 & 5,929 & \textcolor{red}{\ding{55}} & \textcolor{red}{\ding{55}} & \textcolor{ForestGreen}{\ding{51}} & \textcolor{red}{\ding{55}} & \textcolor{red}{\ding{55}} & \textcolor{red}{\ding{55}} & Free-form \\  
        $\text{MMIE}_{\text{math}}$ \cite{mmie} & 26,534 & 20,103 & \textcolor{red}{\ding{55}} & \textcolor{red}{\ding{55}} & \textcolor{ForestGreen}{\ding{51}} & \textcolor{red}{\ding{55}} & \textcolor{red}{\ding{55}} & \textcolor{ForestGreen}{\ding{51}} & Free-form \& Multiple Choice \\  
        Polymath \cite{polymath} & 5,000 & 5,000 & \textcolor{red}{\ding{55}} & \textcolor{ForestGreen}{\ding{51}} & \textcolor{ForestGreen}{\ding{51}} & \textcolor{red}{\ding{55}} & \textcolor{ForestGreen}{\ding{51}} & \textcolor{red}{\ding{55}} & Multiple Choice \\  
        NTSEBench \cite{ntsebench} & 4,642 & 2,728 & \textcolor{red}{\ding{55}} & \textcolor{ForestGreen}{\ding{51}} & \textcolor{red}{\ding{55}} & \textcolor{ForestGreen}{\ding{51}} & \textcolor{ForestGreen}{\ding{51}} & \textcolor{ForestGreen}{\ding{51}} & Multiple Choice \\  
        BSA \footnote{BSA is similar to our single-image setting, but it involves multiple images within a single image, making it a multi-image setting.} \cite{BSA} & 312 & 312 & \textcolor{red}{\ding{55}} & \textcolor{ForestGreen}{\ding{51}} & \textcolor{red}{\ding{55}} & \textcolor{red}{\ding{55}} & \textcolor{red}{\ding{55}} & \textcolor{ForestGreen}{\ding{51}} & Multiple Choice \\  
        MV-MATH \cite{mv_math} & 6,061 & 2,009 & \textcolor{red}{\ding{55}} & \textcolor{ForestGreen}{\ding{51}} & \textcolor{ForestGreen}{\ding{51}} & \textcolor{ForestGreen}{\ding{51}}  & \textcolor{ForestGreen}{\ding{51}}  & \textcolor{ForestGreen}{\ding{51}} & Free-form \& Multiple Choice \\  
        \textbf{Ours} & 6,697 & 1,720 & \textcolor{ForestGreen}{\ding{51}} & \textcolor{ForestGreen}{\ding{51}} & \textcolor{ForestGreen}{\ding{51}} & \textcolor{ForestGreen}{\ding{51}} & \textcolor{ForestGreen}{\ding{51}} & \textcolor{ForestGreen}{\ding{51}} & Multiple Choice \\  
        \bottomrule  
    \end{tabular}  
    }  
\end{table*}

\section{Analysis}

\subsection{Influence of Chain-of-Thought on Model Performance}

Chain-of-thought \cite{cot} reasoning generally enhances model performance, as the Table \ref{tab:results_cot} shows stable improvements across several domains when CoT is enabled. For instance, Qwen-VL-Max exhibits a dramatic 40\% boost in the “Reasoning” task, highlighting the significant impact of structured reasoning on spatial understanding. Gemini2.0-Flash also benefits substantially, with a 15.40 point increase in the “Pattern” category and a 16.66 point rise in “Place” suggesting that CoT particularly aids in tasks requiring complex organizational and geometric reasoning.

%While improvements are evident, the effect of CoT is not uniformly positive across all tasks. GPT-4o, for example, experiences a 20-point drop in “Direction”, indicating that the benefits of chain-of-thought may depend on the model architecture and the nature of the task. Nonetheless, the overall trend supports that incorporating CoT tends to enhance problem-solving abilities, especially in tasks that demand high-level reasoning and pattern recognition.

While improvements are evident, the efficacy of chain-of-thought (CoT) prompting exhibits strong task-dependent variation. CoT consistently enhances performance in multi-step reasoning tasks~(e.g., Pattern and Reasoning tasks), where all models show gains. However, it proves neutral or detrimental in perception-heavy tasks (e.g., Calender and Direction  tasks) due to interference with low-level spatial or temporal processing. Nonetheless, the overall trend supports that incorporating CoT tends to enhance problem-solving abilities, especially in tasks that demand high-level reasoning and pattern recognition.

\begin{figure*}[t]
  \centering
  \includegraphics[width=1.0\textwidth]{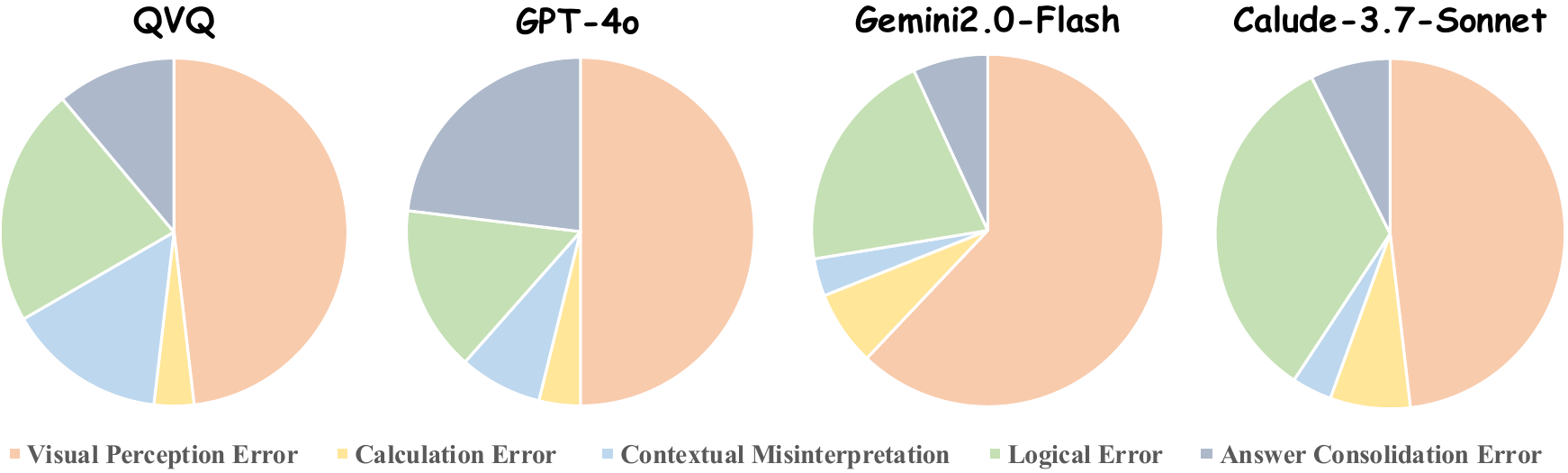}
  \caption{A comparison of error distributions among three model, GPT-4o, Gemini2.0-Flash, and Calude-3.7-Sonnet, across five error categories: visual perception errors, calculation errors, contextual misunderstandings, logical errors, and answer integration errors.}
  \label{fig:error_distribution}
\end{figure*}

\subsection{Comparison with Other Benchmarks}
In comparison to existing visual math benchmarks, our dataset stands out in several important ways as shown in Table \ref{tab:benchmark_compare}. While benchmarks such as Olympiadbench \cite{he2024olympiadbench} and GeoQA \cite{geoqa} focus primarily on specific skills like geometry and logical reasoning, our benchmark includes a broader spectrum of required skills, including temporal, spatial, geometric, logical, and pattern recognition. This comprehensive skill coverage provides a more holistic evaluation of LVLMs. Additionally, our dataset supports multi-image tasks, which is a feature not widely supported by other benchmarks such as Blink \cite{blink} and GeoQA \cite{geoqa}, further enhancing its applicability for real-world tasks that require understanding across multiple visual inputs. Moreover, our benchmark boasts a higher image-question ratio than all other benchmarks, meaning that on average, each question is associated with more images. Finally, our dataset offers multiple-choice answer types for easier evaluation, unlike many other benchmarks that provide free-form answer format which hard to evaluate, such as MM-MATH \cite{mm_math} and U-Math \cite{u_math}.

\subsection{Error Distribution for \textsc{\textsc{VCBench}}}
We define five error types in this benchmark: Visual Perception Error indicates that the model misinterprets or fails to accurately perceive visual content; Calculation Error captures mistakes made during arithmetic computations; Contextual Misinterpretation occurs when the model misreads the textual conditions, such as treating unrelated information as relevant; Logical Error refers to flaws in the reasoning process; and Answer Consolidation Error encompasses failures to directly answer the question or instances where multiple, conflicting answers are provided. 
 We conduct manual error classification for all questions across four top-tier models, enabling precise identification of each model's failure patterns and relative weaknesses across different error categories.
%As shown in Figure ref{fig:error_distribution} Visual Perception Errors are predominant across all models, with Gemini2-Flash exhibiting the highest rate at about 62\%, while QVQ and Claude register lower rates of about 48\%. Calculation Errors remain consistently low (ranging from about 4\% to about 7\%), and Contextual Misinterpretation errors are minimal, particularly for Gemini2-Flash (about 3\%) and Claude (about 4\%), which indicates a relatively robust understanding of textual context. On the other hand, discrepancies are more apparent in the Logical and Answer Consolidation Error rates. Claude shows a significantly high Logical Error rate of about 33\% compared to GPT-4o’s about 15\% and QVQ’s about 22\%, suggesting potential deficiencies in logical reasoning. Moreover, while Answer Consolidation Errors are generally low (QVQ at about 11\% and both Gemini2-Flash and Claude at about 7\%), GPT-4o presents a higher rate of about 23\%, implying that it may occasionally provide unfocused or ambiguous responses.
As shown in Figure~\ref{fig:error_distribution}, Visual Perception Errors are predominant across all models, with Gemini2-Flash exhibiting the highest rate at about 62\%. This persistent pattern across architectures suggests that enhancing visual perception capabilities remains the most critical challenge for multimodal models. Calculation Errors remain consistently low (ranging from about 4\% to about 7\%), indicating that basic arithmetic computation has become relatively robust in modern models. Contextual Misinterpretation errors are minimal, particularly for Gemini2-Flash (about 3\%) and Claude (about 4\%), which indicates a relatively robust understanding of textual context. However, QVQ's comparatively higher rate (6\%) may reflect its tendency toward over-reasoning, where excessive analysis leads to detachment from the original question context.

On the other hand, discrepancies are more apparent in the Logical and Answer Consolidation Error rates. Claude shows a significantly high Logical Error rate of about 33\% compared to GPT-4o’s about 15\% and QVQ’s about 22\%, revealing the weaknesses in its deductive reasoning pipelines. Moreover, while Answer Consolidation Errors are generally low (QVQ at about 11\% and both Gemini2-Flash and Claude at about 7\%), GPT-4o presents a higher rate of about 23\%, suggesting its advanced reasoning capabilities may come at the cost of response discipline, where the model sometimes generates multiple answers rather than a single one. This trade-off between exploratory reasoning and answer precision presents an important optimization target for future iterations.

\subsection{Analysis of Problem Difficulty and Model Performance}
The questions in our benchmark are drawn from existing textbooks and assigned a difficulty coefficient ranging from 0.0 to 1.0, where scores between 0.0 and 0.35 denote easy questions, 0.35 to 0.75 represent medium ones, and 0.75 to 1.0 correspond to hard problems. Interestingly, the results in Table \ref{tab:difficulty_comparison} reveal that questions annotated as hard tend to yield higher accuracy, while the easy and medium problems register lower accuracy. This counterintuitive outcome may be attributed to the fact that simpler questions, which primarily require the identification of patterns rather than intricate computations, pose a different challenge compared to the hard questions that demand complex calculation and structured reasoning.

\section{Conclusion}
This paper introduces \textsc{\textsc{VCBench}}—a comprehensive evaluation framework designed to assess multimodal mathematical reasoning with explicit visual dependency. By addressing the limitations of existing datasets in multi-image integration and cross-modal relational reasoning, our benchmark provides a detailed analysis of 26 state-of-the-art LVLMs across six cognitive domains and 17 task categories. The evaluation reveals significant performance disparities, particularly in areas such as multi-step instruction following, basic visual perception, cross-image consistency, and vulnerability to visual hallucinations.

\section*{Acknowledgement}
The LaTeX template is built upon Meta’s original template.

\bibliographystyle{assets/plainnat}
\bibliography{paper}

\newpage

\appendix

\section{Appendix}
\label{sec:appendix}

\subsection{Experiment Details}

\begin{table*}[th!]
\caption{Generation parameters for LVLMs (with grouped configurations).}
\small
\centering
\begin{tabular}{p{0.24\linewidth} | p{0.69\linewidth}}
\toprule
\textbf{Model} & \textbf{Generation Setup} \\
\midrule
\multirow{2}{*}{GPT-4o-mini \& GPT-4o} & API URL: \url{https://api.openai.com/v1/chat/completions}
temperature = 0.2, max\_tokens = 1024 \\
\midrule
Claude-3.7-Sonnet & API URL: \url{https://api.anthropic.com/v1/messages}, temperature = 0.2, max\_tokens = 1024 \\
\midrule
Gemini2.0-Flash & API URL: \url{https://generativelanguage.googleapis.com/v1beta/models/gemini-pro:generateContent}, temperature = 0.2, max\_tokens = 1024 \\
\midrule
Qwen-VL-Max & Use dashscope package, temperature = 0.2, max\_new\_tokens = 1024 \\
\midrule
\multirow{14}{*}{Open-Source Models} & \textit{Same parameters for all below:} \\
& Deployed by vllm, with do\_sample = True, temperature = 0.2, max\_new\_tokens = 1024 \\
& • Idefics3-8B \\
& • LLaMA-3.2-90B-Vision-Instruct \\
& • Emu2-Chat \\
& • DeepSeek-VL2 \\
& • Mantis-CLIP \\
& • LLaVA-Interleave-7B \\
& • Phi-3.5-vision-instruct \\
& • InternVL-2.5 \\
& • LLaVA-OneVision-7B/72B \\
& • Gemma3-27B-it \\
& • Mistral-Small-3.1-24B-Instruct \\
& • Qwen2.5-VL-7B/72B-Instruct \\
\midrule
QVQ-72B-Preview & do\_sample = True, temperature = 0.2, max\_new\_tokens = 2048 \\
\midrule
G-LLaVA-7B/13B & do\_sample = True, temperature = 0.2, max\_new\_tokens = 1024 \\
\midrule
MathLlava & do\_sample = True, temperature = 0.2, max\_new\_tokens = 1024 \\
\bottomrule
\end{tabular}
\label{tab:experiment_details}
\end{table*}

%\subsection{Additional Dataset Analysis}

\subsection{Case Studies}
\begin{figure*}
  \centering
  \includegraphics[width=0.88\textwidth]{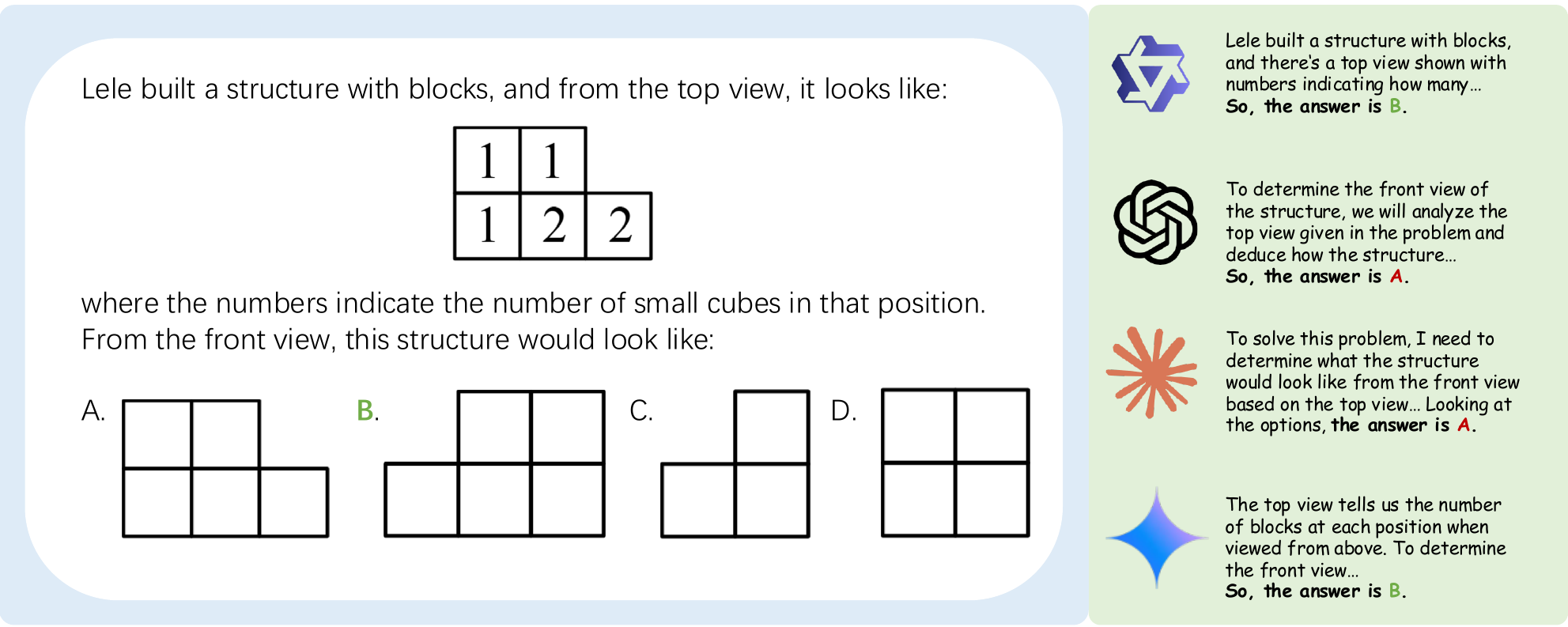}
  \caption{Case for Visual Perception Error.}
  \label{fig:casestudy1}
\end{figure*}

\begin{figure*}
  \centering
  \includegraphics[width=0.88\textwidth]{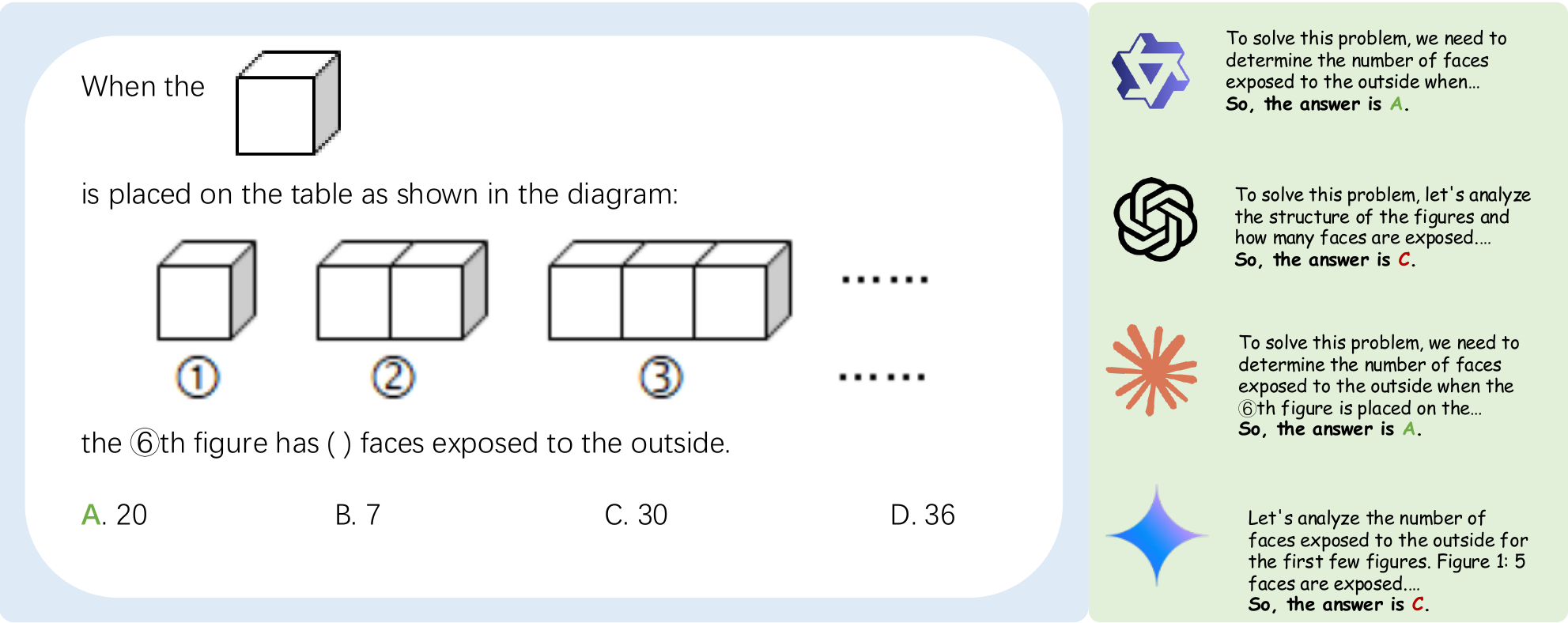}
  \caption{Case for Calculation Error.}
  \label{fig:casestudy2}
\end{figure*}

\begin{figure*}
  \centering
  \includegraphics[width=0.88\textwidth]{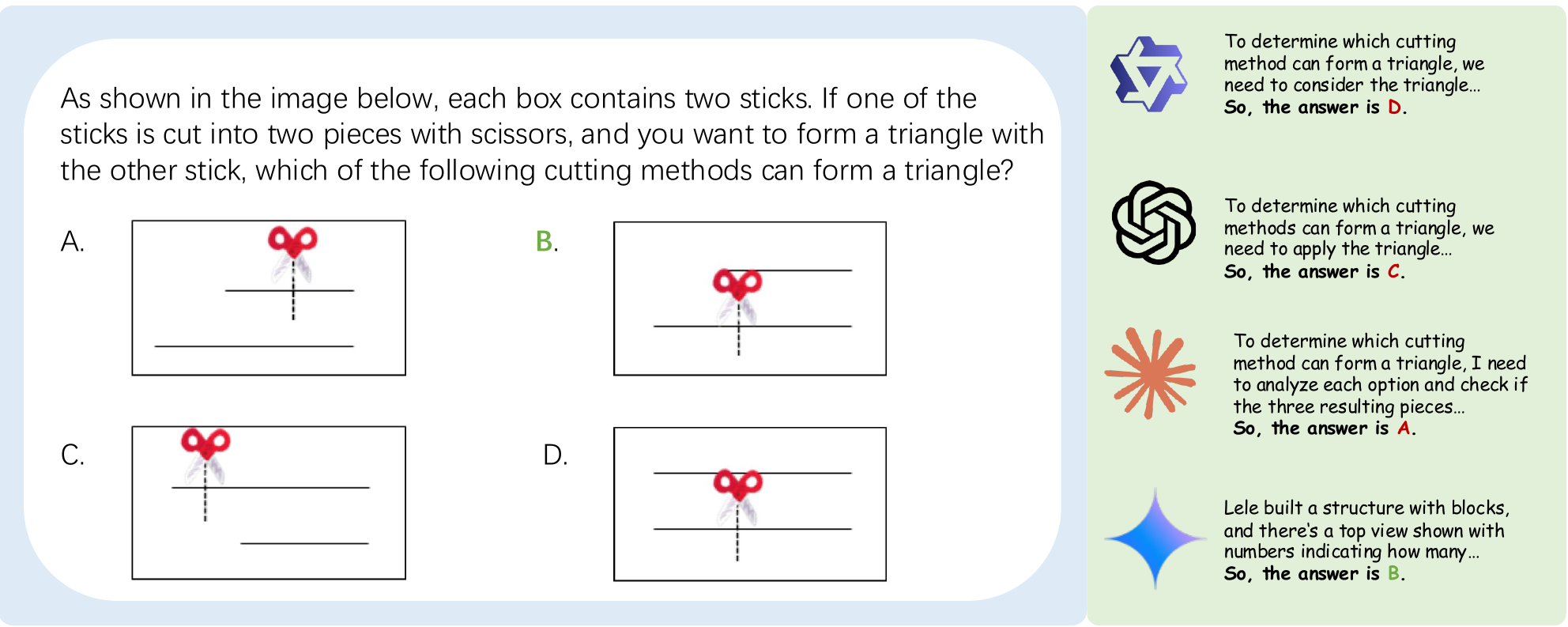}
  \caption{Case for Contextual Misinterpretation.}
  \label{fig:casestudy3}
\end{figure*}

\begin{figure*}
  \centering
  \includegraphics[width=0.88\textwidth]{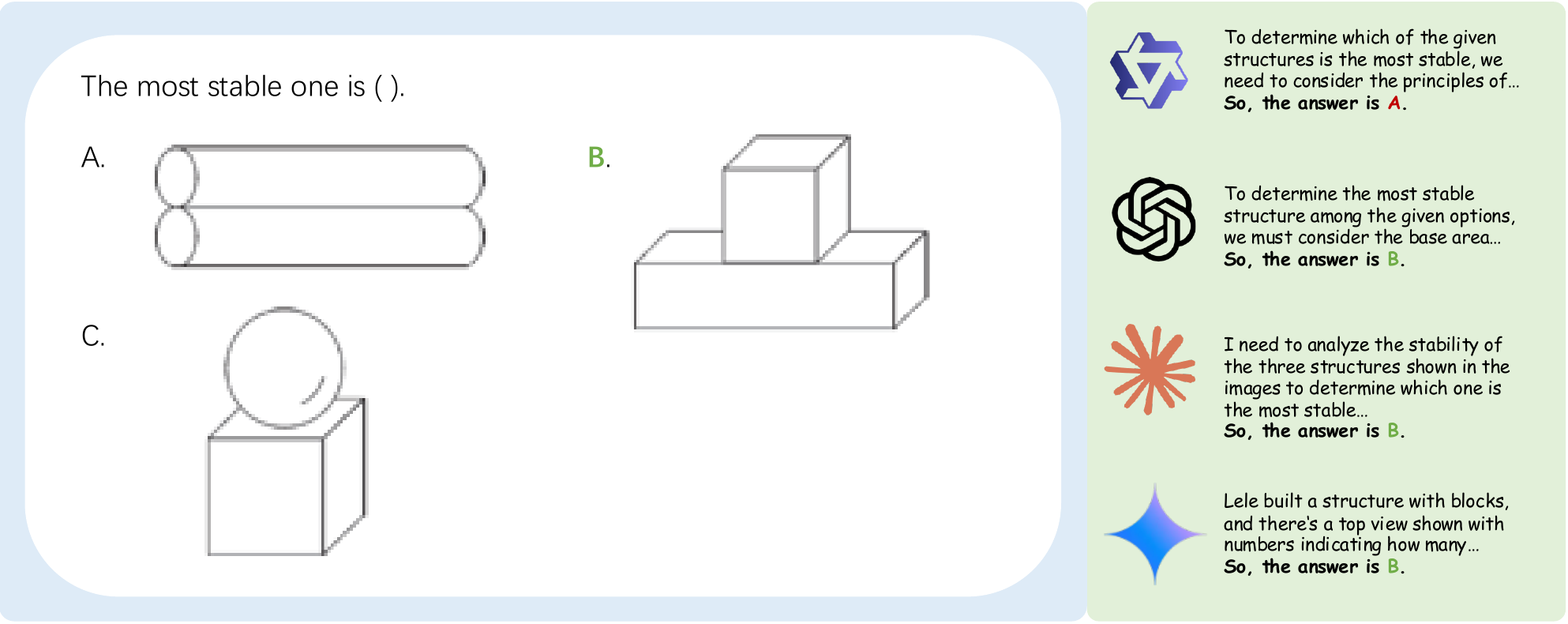}
  \caption{Case for Logical Error.}
  \label{fig:casestudy4}
\end{figure*}

\begin{figure*}
  \centering
  \includegraphics[width=0.88\textwidth]{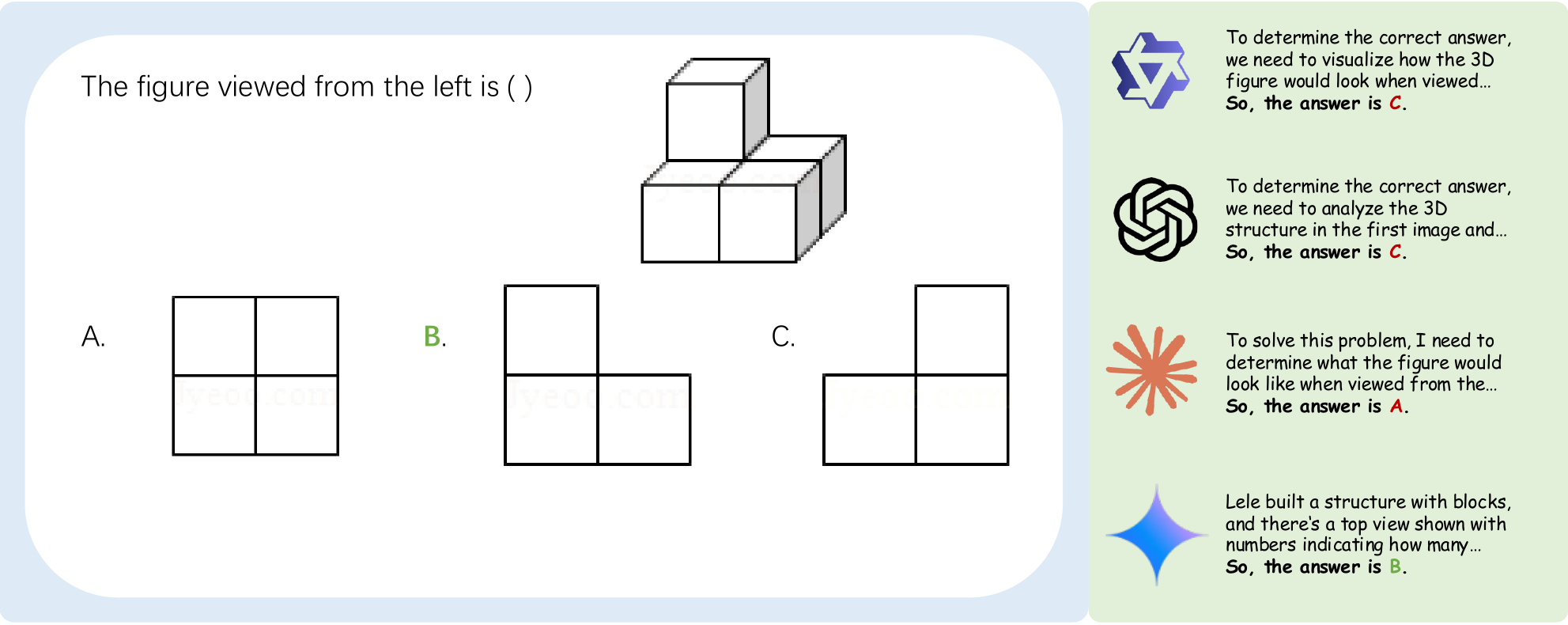}
  \caption{Case for Answer Consolidation Error.}
  \label{fig:casestudy5}
\end{figure*}

\subsection{Prompt for Experiment}

\begin{table*}[h]
    \centering
    \caption{Inference Prompt.}
    \begin{minipage}{\textwidth}
        \centering
        \begin{tcolorbox}[title=Inference Prompt]
You are a helpful AI assistant.

Please answer the following questions and output the answer options directly.

\textit{Question: \{ question \}}

        \end{tcolorbox}
    \end{minipage}
    \label{tab:inference_prompt}
\end{table*}

\begin{table*}[h]
    \centering
    \caption{Inference Prompt with Chain-of-Thought.}
    \begin{minipage}{\textwidth}
        \centering
        \begin{tcolorbox}[title=Inference Prompt with Chain-of-Thought]
You are a helpful AI assistant.

Please think step by step before answer the following questions and the output the answer.

\textit{Question: \{ question \}}

        \end{tcolorbox}
    \end{minipage}
    
    \label{tab:inference_prompt_cot}
\end{table*}

\begin{table*}[h]
    \centering
    \caption{LLM-Based Evaluation Prompt.}
    \begin{minipage}{\textwidth}
        \centering
        \begin{tcolorbox}[title=LLM-Evaluation Prompt]
You are an answer evaluator. I will give you a response and an answer.

Please tell me whether this response is correct or wrong. Just answer \textbf{yes} or \textbf{no}.
    
For example,

Response: The figure that cannot be folded into a cube is: C. <image>

Correct Answer: B

So, you need to respond \textbf{no} only.

Response: The unfolded shape of the cube is: B. <image>

Correct Answer: B

So, you need to respond \textbf{yes} only.

Here is the response and correct answer I want you to evaluate.

\textit{Response: \{ model response \}}

\textit{Correct Answer: \{ correct answer \}}

        \end{tcolorbox}
    \end{minipage}
    
    \label{tab:evaluation_prompt}
\end{table*}

\end{document}